\documentclass[11pt,draftclsnofoot,onecolumn,journal]{IEEEtran}
\usepackage{amsmath,amsfonts}
\usepackage{algorithmic}
\usepackage{algorithm}
\usepackage{array}
\usepackage[caption=false,font=normalsize,labelfont=sf,textfont=sf]{subfig}
\usepackage{textcomp}
\usepackage{stfloats}
\usepackage{url}
\usepackage{verbatim}
\usepackage{graphicx}
\usepackage{cite}
\usepackage[backref=page]{hyperref}
\usepackage{booktabs}
\usepackage[para,online,flushleft]{threeparttable}

\usepackage{amsthm,amssymb,amscd,mathrsfs}
\usepackage{enumitem}

\usepackage[capitalize,noabbrev]{cleveref}

\usepackage{wrapfig}

\usepackage{crossreftools}
\usepackage{xargs} 
\usepackage{xstring}
\usepackage{etoolbox}

\usepackage{etoc}
\usepackage[bibliography=common]{apxproof}

\newtheoremrep{thm}{Theorem}
\newtheoremrep{cor}[thm]{Corollary}
\newtheoremrep{lemma}[thm]{Lemma}
\newtheoremrep{fact}[thm]{Fact}
\newtheoremrep{prop}[thm]{Proposition}

\newtheorem{definition}[thm]{Definition}

\theoremstyle{remark}

\crefname{thm}{theorem}{theorems}
\crefname{assumption}{assumption}{assumptions}
\crefname{cor}{corollary}{corollaries}
\crefname{prop}{proposition}{propositions}
\crefname{lemma}{lemma}{lemmas}
\crefname{fact}{fact}{facts}

\usepackage{mdframed}
\newmdtheoremenv{algo}{Algorithm}
\newmdtheoremenv{procedure}{Decision process}

\newlist{assnum}{enumerate}{1} %
\setlist[assnum]{label=(\roman*), ref=\theassumption(\roman*)}
\crefalias{assnumi}{assumption} 

\newlist{lemnum}{enumerate}{1} %
\setlist[lemnum]{label=(\roman*), ref=\thelemma(\roman*)}
\crefalias{lemnumi}{lemma} 

\newlist{thmnum}{enumerate}{1} %
\setlist[thmnum]{label=(\roman*), ref=\thethm(\roman*)}
\crefalias{thmnumi}{thm} 

\newlist{cornum}{enumerate}{1} %
\setlist[cornum]{label=(\roman*), ref=\thecor(\roman*)}
\crefalias{cornumi}{cor} 

\newlist{definitionnum}{enumerate}{1} %
\setlist[definitionnum]{label=(\roman*), ref=\thedefinition(\roman*)}
\crefalias{definitionnumi}{definition} 

\newlist{propnum}{enumerate}{1} %
\setlist[propnum]{label=(\roman*), ref=\theproposition(\roman*)}
\crefalias{propnumi}{prop}

\newlist{examplenum}{enumerate}{1} %
\setlist[examplenum]{label=(\roman*), ref=\theexample(\roman*)}
\crefalias{examplenumi}{example}

\usepackage{dsfont}
\usepackage{nicefrac}

\DeclareMathOperator*{\argmax}{arg\,max}
\DeclareMathOperator*{\argmin}{arg\,min}

\newcommand{\R}{\mathbb{R}}

\newcommand{\lmo}{\operatorname{lmo}}
\newcommand{\sign}{\operatorname{sign}}
\newcommand{\RMS}{\mathrm{RMS}}

\newcommand{\norm}[1]{\left\Vert{#1}\right\Vert}

\usepackage{tcolorbox}
\newtcolorbox{defbox}{colback=black!5!white,colframe=black!75!black}
\newtcolorbox{asmbox}{colback=black!5!white,colframe=black!75!black}
\newtcolorbox{thmbox}{colback=red!5!white,colframe=red!75!black}

\usepackage{braket}

\newcommand{\diag}{\operatorname{diag}}

\newcommandx{\QC}[2][1={},2={}]{\ifstrempty{#1}{Q#2}{Q_{#1}\ifstrempty{#2}{}{(#2)}}}
\newcommandx{\PC}[2][1={},2={}]{\ifstrempty{#1}{P#2}{P_{#1}\ifstrempty{#2}{}{(#2)}}}
\newcommandx{\HC}[2][1={},2={}]{\ifstrempty{#1}{H#2}{H_{#1}\ifstrempty{#2}{}{(#2)}}}
\newcommandx{\MC}[2][1={},2={}]{\ifstrempty{#1}{M#2}{M_{#1}\ifstrempty{#2}{}{(#2)}}}

\newcommandx{\EF}[2][1={k},2={}]{\mathbb E\ifstrempty{#1}{}{_{#1}}#2}

\newcommand{\blue}[1]{{\color[rgb]{0,0,1}#1}}

\usepackage{adjustbox}

\usepackage{array,multirow,hhline,graphicx}
\usepackage{colortbl}
\newcommand{\STAB}[1]{\begin{tabular}{@{}c@{}}#1\end{tabular}}

\usepackage{pifont}

\newcommand\tp[1]{{#1}}

\makeatletter %
\renewcommand{\subsubsection}{\@startsection{subsubsection}{3}{\z@}%
  {0ex}{0ex}{\normalfont\bfseries}}
\makeatother
\renewcommand{\appendix}{} %

\begin{document}

\title{Training Neural Networks at Any Scale}

\author{Thomas Pethick$^*$\thanks{$^*$École Polytechnique Fédérale de Lausanne (EPFL), Laboratory for Information and Inference Systems (LIONS)}, Kimon Antonakopoulos$^*$, Antonio Silveti-Falls$^\dagger$\thanks{$^\dagger$CVN, CentraleSupélec, Université Paris-Saclay, Inria},\\  Leena Chennuru Vankadara$^\ddagger$ \thanks{$^\ddagger$Gatsby Computational Neuroscience Unit, University College London}, Volkan Cevher$^*$}

\markboth{}%
{Aviyente \MakeLowercase{\textit{et al.}}: Author Guidelines for Special Issue Articles of IEEE SPM}

\maketitle

\vspace{-15mm}
\section*{Introduction}\label{sec:outline}
Deep learning thrives on the powerful interplay between data, architectures, and optimization. As the sizes of foundation models continuously grow in scale with 405B Llama models and 671B DeepSeek models, we see unprecedented performance and new transformative capabilities. However, this increase in scale also demands optimization methods that scale efficiently and reliably.

\looseness=-1 Traditionally, optimization theory for neural networks has treated training as a black-box optimization problem. By black-box optimization, we mean optimizing without taking into account possible structure present in the objective function or its gradients. 
While this generality has been convenient, it has limited our ability to exploit inherent characteristics of the problem that can significantly enhance performance. Recent advances \cite{pethick2025trainingdeeplearningmodels,jordan2024muon,bernstein2024modular}, however, have demonstrated the power of jointly designing neural network architectures, initialization schemes, and optimization algorithms, while understanding their training regimes through the notion of feature learning.

\looseness=-1Key insights have emerged from the study of infinite-width neural network limits, particularly through frameworks such as Neural Tangent Kernel (NTK) \cite{jacot2018neural} and mean-field analyses \cite{chizat}. A landmark development, called the Maximal Update Parameterization ($\mu$P) \cite{yang2021tensor}, showed that feature learning can be ensured when the network architecture, initialization, and layerwise stepsizes are carefully aligned. Remarkably, this alignment also results in a beneficial property called hyperparameter transfer, in which hyperparameter settings (like learning rate) remain optimal across model scales, significantly stabilizing the training.

Furthermore, optimization methods have recently evolved to be both more reliable and faster by explicitly leveraging the network's structure as a collection of layers. A notable example is the Shampoo optimizer \cite{Gupta2018ShampooPS}, which topped the recent AlgoPerf benchmark challenge hosted by MLCommons. Shampoo explicitly utilizes the network's structure as a collection of weight matrices by preconditioning the update to each layer's weight matrix separately, connecting elegantly to spectral descent methods dating back to \cite{carlson2015stochastic}. As we explore in this article, the spectral geometry emerges naturally from analyzing how signals or features propagate through the network, and can be used for more effective training.

To this end, we present a contemporary perspective on deep learning optimization, highlighting key connections between optimization techniques and neural network architectures. Our goal is to equip both experts and practitioners 
with actionable insights on choosing and configuring optimizers in tandem with architectures to achieve scale-agnostic neural network training.

\looseness=-1\paragraph*{Outline}
We begin by introducing a general framework for deriving optimization algorithms that leverage problem geometry through an appropriate choice of norm. Perhaps unexpectedly, we find that the classical stochastic Frank–Wolfe method for constrained optimization \cite{mokhtari2020stochastic} emerges as a central component of this framework.

Second, we examine scaling rules: \textit{how to optimally adjust the training recipe as the problem size varies}. We analyze signal propagation in neural networks under scaling, focusing on how to balance training stability with meaningful feature evolution, particularly as network width increases.

\begin{figure*}[t]
    \centering
    \includegraphics[width=0.30225\textwidth]{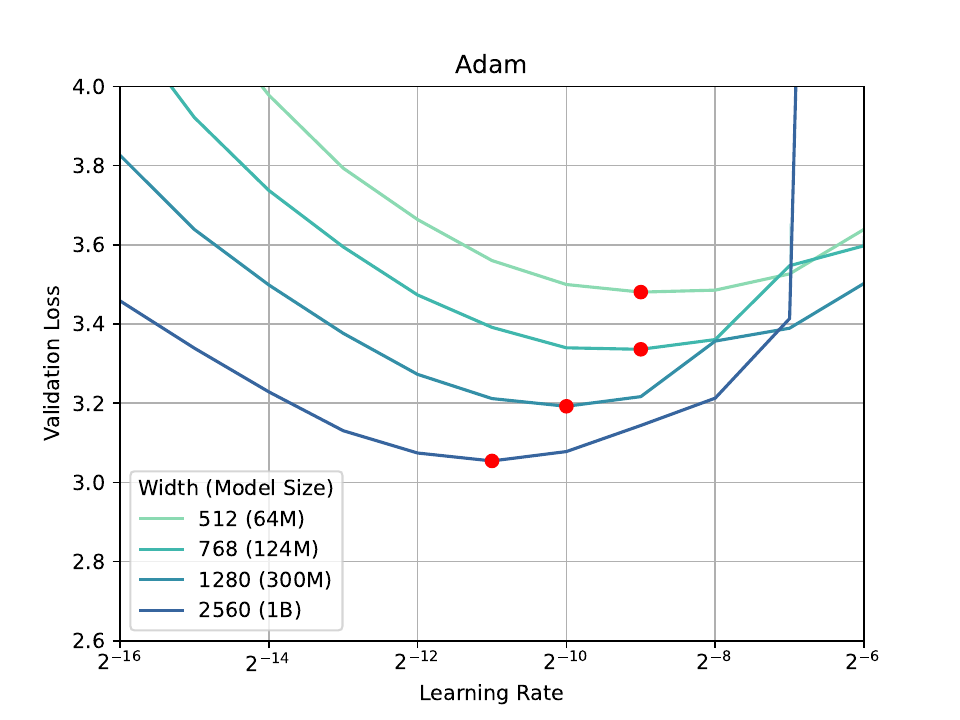}
    \includegraphics[width=0.30225\textwidth]{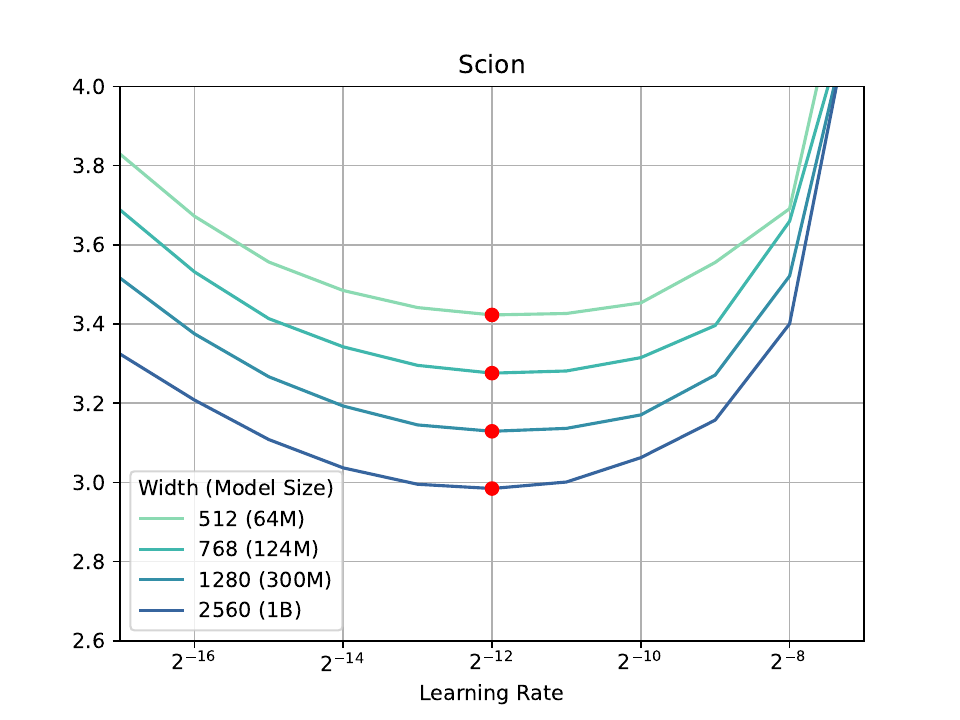}
    \includegraphics[width=0.3775\textwidth]{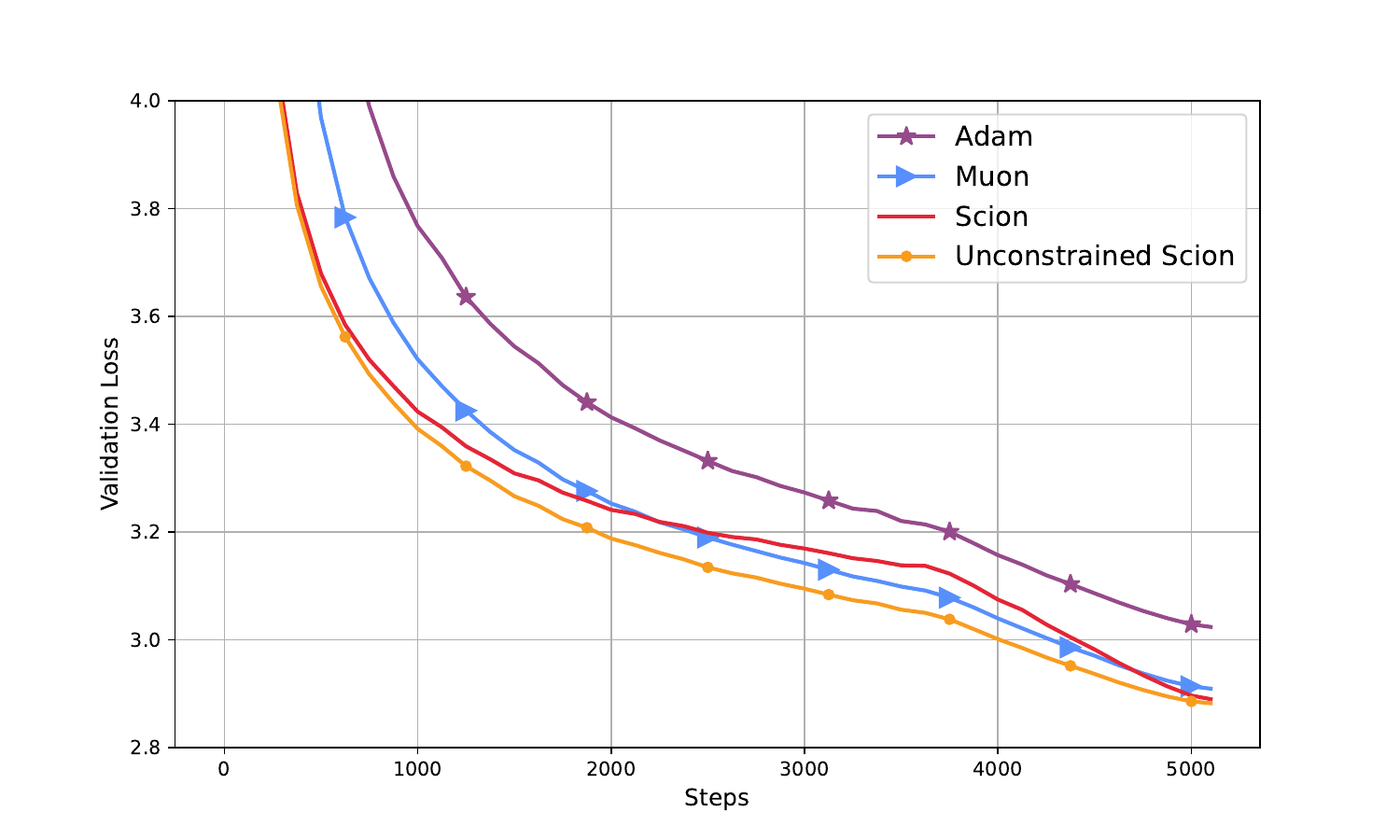}
    \caption{
    Different optimizers for training large language models (LLMs), illustrated with a GPT model having up to 3 billion parameters (plots from \cite{pethick2025trainingdeeplearningmodels}).
    A modern optimizer like Scion (middle) exploits the geometry of the neural network to achieve faster convergence (right), while simultaneously being parameterized such that the optimizer becomes agnostic to scale (middle, all other parameters constant). This design leads to the transfer of hyperparameters, like learning rate, across model width, as demonstrated above. Performance is also boosted in large-scale training (3 billion parameters, right).
    } \label{fig:GPT}
\end{figure*}

Neural networks can learn features even in the infinite-width limit when a specific \textit{parameterization}, i.e., a {layerwise scaling} of initialization variances and learning rates, is employed. Under common parameterizations, such as the Neural Tangent Parameterization (NTP) \cite{jacot2018neural}, infinitely wide networks operate in the so-called lazy regime and do not exhibit feature learning. In contrast, when the network satisfies the maximal update criterion \cite{yang2021tensor}, which promotes maximal feature learning across all layers, we empirically observe hyperparameter transfer (\emph{cf.}, \Cref{fig:GPT}). In practice, this means that the optimal learning rate becomes independent of model width and a learning rate tuned on a smaller model can be directly applied to a larger one. We conclude with a discussion of current limitations and open challenges.

\section*{Geometry-Aware Optimization Methods}\label{sec:scope}
\vspace{-2mm}

\looseness=-1Adam, an adaptive optimizer based on prior methods like AdaGrad and RMSProp, has dominated deep learning by dynamically changing the geometry in which the updates to parameters are computed. Methods like this adapt to the local geometry on-the-fly using the gradients seen by the algorithm so far. This enables reliable training with little tuning. However, these algorithms also treat the neural network (NN) to be trained as a black-box with all parameters treated irrespective of their role or position in the architecture. 

A different approach instead chooses an appropriate geometry to compute the updates to parameters  \emph{prior} to the optimization phase. This choice of geometry is encoded through a norm on the parameters and is deliberately made with the NN architecture to be trained in mind, in contrast to the previously described adaptive methods. In the following section, we lay out both of these algorithmic approaches under a unified algorithmic template involving steepest descent and conditional gradient methods.

\vspace{-4mm}
\subsection*{Preliminaries}\vspace{-1mm}

We begin by describing the mathematical formalism that we will use to discuss different algorithms and initializations. We frame training as solving the following non-convex optimization problem:
\begin{equation}\label{eq:min}
\min_{x \in \mathcal X} f(x),
\end{equation}
where $f$ is $L$-smooth (i.e., its gradient $\nabla f$ is Lipschitz continuous with constant $L>0$ in some general norm)
and the domain $\mathcal{X}$ is unconstrained and given by $\mathbb{R}^p$, where $p$ is the number of NN parameters. %

We assume access to some stochastic oracle $f(\cdot,\xi)$ instead of $f(\cdot)$ directly, where $\xi$ is a random variable, typically representing a minibatch of data. From an algorithmic perspective, we aim to solve \eqref{eq:min} with iterative methods that use a \emph{stochastic first-order oracle} (SFO).
At each iteration, the algorithm can query a black-box mechanism at a given point $x$ that returns a stochastic estimate $g(x,\xi)$ of $\nabla f(x)$. 

For instance, the workhorse of modern deep learning, AdamW \cite{loshchilov2017decoupled}, presented below \emph{without the bias correction} proceeds with the following  updates   at each iteration $k$, given $\beta_{1,2} \in [0,1)$:
$$
d^k = \beta_1 d^{k-1} + (1-\beta_1) g^k, \
v^k = \beta_2 v^{k-1} + (1-\beta_2) (g^k \odot g^k), \
x^{k+1} = (1- \underbrace{s_k\lambda}_{\color{blue} =\lambda_k})  x^k - \underbrace{s_k\gamma }_{\color{blue}=\gamma_k} \frac{d^k}{\sqrt{v^k+\varepsilon^2}},
$$
 where multiplication $\odot$, division and square root are taken coordinatewise; $\lambda \in[0,1)$ is the decoupled weight decay constant; $\gamma$ is the stepsize; $s_k$ is the learning rate scheduler; and $\varepsilon>0$  is a small numerical constant. We take $d^0=g^0$ and $v^0 = g^0 \odot g^0$  to avoid the bias correction.  
In the sequel, we refer to $\lambda_k$ and $\gamma_k$ as the scheduled weight decay and scheduled stepsize respectively.
 
 Intuitively, AdamW builds a diagonal preconditioner \emph{on-the-fly} that rescales coordinates by an exponential moving average (EMA) of their gradient energy via $\beta_2$. As opposed to directly using the stochastic gradients $g^k$, AdamW uses their (Polyak) momentum accumulation via $\beta_1$. AdamW relies on the weight decay parameter $\lambda>0$ to prevent the parameters from growing unbounded, which is observed to improve generalization. 

\looseness=-1To connect AdamW to a broader picture, we introduce the concept of ``dual feedback,''  denoted by $d^k$, which represents a key signal the optimizer uses at each iteration. In the deterministic setting, $d^k$ can simply be the gradient $\nabla f(x^k)$ while in the stochastic setting choosing $d^k$ can be more sophisticated, involving things like variance reduction. For now we will consider two main examples, in which $d^k$ is taken to be $g^k$ directly or a momentum accumulation of previous $g^k$; both can be defined recursively via $d^k = (1-\alpha_k)d^{k-1} + \alpha_kg^k$ for some $\alpha_k\in(0,1]$, which is equivalent to the estimator in Adam when $\alpha_k = 1-\beta_1$.
This momentum estimator reduces variance but is a biased estimate of $\nabla f(x^k)$.

Having established the dual feedback, we  now introduce our general master algorithmic template: %
\begin{equation}
\label{eq:gen_form}
\textstyle     x^{k+1}\in \argmin_{x\in \mathcal{X}} \gamma_k\langle d^{k}, x\rangle+h_k(x), 
\end{equation}
where $\gamma_k>0$ is the stepsize and $h_k$ is a convex function which facilitates a tractable solution of \eqref{eq:gen_form}, i.e., a \emph{regularizer}. The generality of this formulation allows us to unify several celebrated algorithms for training neural networks within the somewhat classical framework below with different choices of $d^k$ and $h_k$. 

In what follows, we explore the flexibility of the scheme in \eqref{eq:gen_form} to adapt to the geometry of \eqref{eq:min}, both on-the-fly or a priori, through appropriate choice of the regularizer $h_k$. We will also discuss how the two other parts, the dual feedback $d^k$ and the stepsize $\gamma_k$, are essential for treating the stochastic case.

\subsection*{On-the-fly adaptation to the geometry}
We start with describing how AdamW adapts on-the-fly to the problem geometry. To make this explicit, we first consider the norm associated to the Mahalanobis inner product in a Euclidean space as follows:  $\norm{x}_{2,H}=\sqrt{\langle x,H^{1/2}x \rangle},$ where $H\succ 0$ is a positive define matrix. Then, given a sequence of such matrices $\{H_k\}$, we can define a sequence of regularizers that will play a unifying role for adaptation: 
\begin{equation}\label{eq:regularizers}
    h_k(x)=\tfrac{1}{2}\|x-(1-\lambda_k)x^k\|_{2,H_k}^2.
\end{equation}

Using this family of regularizers, \eqref{eq:gen_form} reduces to the preconditioned stochastic gradient descent  method with decoupled weight decay \ref{eq:PreSGD}, given by the following generic template: 
\begin{equation*}\label{eq:PreSGD}
\tag{PreSGDW}
x^{k+1} = (1-\lambda_k) x^k - \gamma_k H_k^{-1/2} d^k.
\end{equation*}
Alternatively, choosing $d^k=g^k$, $H_k = I$ (identity), and weight decay parameter $\lambda=0$, we reach the stochastic gradient descent (SGD). By choosing $d^k$ as the momentum estimator with $\alpha_k=1-\beta_1$ and $H_k= \text{diag}(v^k+\varepsilon^2 I)$, we reach AdamW.

\begin{table}[!t]
\centering
\caption{A chronological taxonomy of selected first-order optimization methods under \ref{eq:PreSGD}}
\label{table:taxanomy}
\scalebox{0.97}{
\begin{threeparttable}
\begin{tabular}{|l|c|c|c|}
\hline
\textbf{Algorithm} & \textbf{Reference} & \textbf{Dual feedback $d^k$} & \textbf{Preconditioner $H_k$}\tnote{1} \\
\hline
{SGD} & \cite{Polyak1987}
& $g^k$ 
& $[H_k]_{ii} = 1;  \forall  i \in[p]$ \\
\hline
{Momentum (Polyak)} & \cite{Polyak1987} 
& $\beta_1 d^{k-1}+(1-\beta_1) g^k$ 
& $[H_k]_{ii} = 1;  \forall  i \in[p]$  \\
\hline
{AdaGrad (diagonal)} &  \cite{duchi2011adaptive,mcmahan2010adaptive}
& $g^k$ 
& $[H_k]_{ii} = [H_{k-1}]_{ii}+ (g_i^k)^2; \forall i \in[p]$ \\
\hline
{AdaGrad (full)} &  \cite{duchi2011adaptive,mcmahan2010adaptive}
& $g^k$ 
& $H_k = H_{k-1} +  g_i^k (g_i^k)^\top$ \\
\hline
{RMSProp} &   \cite{hinton2012neural}
& $g^k$ 
& $[H_k]_{ii} = \beta_2[H_{k-1}]_{ii} + (1-\beta_2)(g_i^k)^2; \forall i \in[p]$  \\
\hline
{Adam / AdamW} &  \cite{kingma2014adam,loshchilov2017decoupled}
& $\beta_1 d^{k-1}+(1-\beta_1) g^k$ 
& $[H_k]_{ii} = \beta_2[H_{k-1}]_{ii} + (1-\beta_2)(g_i^k)^2; \forall i \in[p]$  \\
\hline
Stochastic $\ell_\infty$-descent\tnote{2} &   \cite{carlson2015stochastic,carlson2016stochastic}
& $g^k$ 
& $[H_k]_{ii} =  (g_i^k)^2; \forall i \in[p]$ \\
\hline
\shortstack{Stochastic spectral descent}\tnote{3} &   \multirow{2}{*}{\cite{carlson2015stochastic,carlson2016stochastic}}
& $G_k= \text{mat}(g^k)$ 
& $R_k\otimes I$; $I \otimes L_k$; or $R_k^{1/2}\otimes L_k^{1/2}:\ \begin{cases}
R_k= G_k G_k^\top\\
L_k= G_k^\top G_k
\end{cases}$ \\
\hline

SignSGD / Signum  &   \cite{bernstein2018signsgd}
& $g^k / \beta_1 d^{k-1}+(1-\beta_1) g^k$ 
& $[H_k]_{ii} = (g_i^k)^2; \forall i \in[p]$ \\
\hline
{Shampoo} &  \multirow{2}{*}{\cite{Gupta2018ShampooPS}}
& $G_k= \text{mat}(g^k)$ 
&  $R_k^{1/2}\otimes L_k^{1/2}:\ \begin{cases}
R_k= \beta_2 R_{k-1} + (1-\beta_2) G_k G_k^\top\\
L_k= \beta_2 L_{k-1} + (1-\beta_2) G_k^\top G_k
\end{cases}$ \\
\hline
uScion / {Muon} (spectral) &  \multirow{2}{*}{ \cite{jordan2024muon,pethick2025trainingdeeplearningmodels}}
& $\beta_1 d^{k-1}+(1-\beta_1) G_k$ 
& $R_k\otimes I$; $I \otimes L_k$; or $R_k^{1/2}\otimes L_k^{1/2}:\ \begin{cases}
R_k=  G_k G_k^\top\\
L_k=  G_k^\top G_k
\end{cases}$ \\
\hline
\end{tabular}
\begin{tablenotes}
\item[1] Unless stated otherwise, we have $[H_k]_{ij} = 0;  \forall i\ne j \in[p]$ ;  \item[2] The stepsize $\gamma_k$ needs to be scaled by $\|d^k\|_1$. \\\item[3] The stepsize $\gamma_k$ needs to be scaled by $\|d^k\|_{S_1}$. 
\end{tablenotes}
\end{threeparttable}
}
\end{table}

Table \ref{table:taxanomy} summarizes several popular first-order optimization algorithms as specific instantiations of \ref{eq:PreSGD}. Each method differs only in two essential modeling choices:
(i) how the dual feedback $d^k$ is formed from the observed gradients, and
(ii) what geometry, encoded via the \emph{preconditioner} $H_k$, and hence the regularizer $h_k$ above, is used to compute the next iterate.
These differences reveal how existing optimizers are related in the way they adapt  to the geometry of the underlying problem. 

\looseness=-1Methods like SGD and Polyak momentum operate under a fixed Euclidean geometry with an identity preconditioner. In contrast, AdaGrad, RMSProp, and AdamW introduce a diagonal preconditioner that evolves with the observed gradients, endowing the parameter space with an anisotropic geometry that varies across iterations. From the table, it is clear that RMSProp is a generalization of diagonal AdaGrad with a tunable $\beta_2$ parameter in the computation of $H_k$, while Stochastic $\ell_\infty$-descent is the limit of RMSProp with $\beta_2\rightarrow 0$. AdamW further expands this class of algorithms with momentum-based dual feedback, reducing dependence on (unbiased) minibatch stochastic gradients with $\beta_1$.

\looseness=-1 In general, going beyond diagonal preconditioners in \ref{eq:PreSGD} can dramatically improve  performance but this is especially resource heavy in deep learning, including memory and compute. However, we can obtain a practical compromise by maintaining $\{H_k = R_k \otimes L_k\}$ with kronecker-factors $\{R_k\}$ and $\{L_k\}$ for right and left preconditioning, respectively. We can then treat the stochastic first-order dual feedback $d^k$ layerwise as a matrix $D_k = \text{mat}(d^k)$ and rely on the identity $H_k^{-1/2}d^k = \text{vec}\left(L_k^{-1/2}D_kR_k^{-1/2}\right)$. Table \ref{table:taxanomy} describes Stochastic spectral descent (SSD) and Shampoo as two algorithms that use this perspective. We see that SSD is the limit of Shampoo, with both forming the spectral analogs of stochastic $\ell_\infty$-descent and AdaGrad, respectively.

It is worth noting that the inclusion of the weight decay term $\lambda_k$ inside the regularizer \eqref{eq:regularizers} is ad-hoc; we write things this way just so that all methods fit the same mold in Table~\ref{table:taxanomy}. Indeed, it is introduced primarily to obtain the generic form \ref{eq:PreSGD} that unifies preconditioning and decoupled weight decay within a single framework. In the sequel, we shift focus to a priori geometry design, where we will encode architectural knowledge directly into the optimizer through a norm that does not change across iterations of the algorithm. Also within this new framework, the weight decay $\lambda$ will play a more natural and prominent role.

\subsection*{A priori adaptation to the geometry}

When training NNs, the form of \eqref{eq:min} has a particular, predictable structure induced by the network architecture and the input domain. This useful yet relatively generic information can be encoded geometrically via the construction of a norm on $\mathcal{X}$. The norm $\|\cdot\|$, together with its dual norm,
\begin{equation}\tag{Dual Norm}
    \|u\|_{\ast}:=\max_{\|x\|\leq 1}\langle u,x\rangle,
\end{equation}
can guide the choice of $h_k$ to get a priori adaptation of the optimizer to the new geometry.
We will describe how to construct such a norm without an inner product structure in later sections; for now we start by introducing two prominent families of regularizers based on this non-Euclidean paradigm. %

\subsubsection*{Steepest descent}

One natural way to generalize \ref{eq:SGD} to a new geometry is to simply regularize with a non-Euclidean norm $h_k(x)=\tfrac{1}{2}\|x-x^k\|^2$.
With this choice, the algorithmic template \eqref{eq:gen_form} reduces to the stochastic steepest descent (SD) method  in a normed space \cite{carlson2015stochastic,carlson2016stochastic}, which can be expressed concisely by defining the sharp-operator, $s^\sharp \in \argmax_{x \in \mathcal X} \{ \braket{s,x} - \tfrac{1}{2}\|x\|^2 \}$, introduced in \cite{nesterov2012efficiency}:
\begin{equation*}\label{eq:SSD}
\tag{Stochastic SD}
x^{k+1} = x^k - \gamma_k [d^k]^\sharp.
\end{equation*}
Here, the sharp-operator maps the dual feedback $d^k$ to the primal space according to the chosen geometry.

Key for deep learning applications are geometries that induce so-called \emph{dense} updates.
An illustrative example is given by the $\ell_\infty$-vector norm, for which \ref{eq:SSD} reduces to a sign-based update, e.g., Stochastic $\ell_\infty$-descent (\textit{cf.}, Table \ref{table:taxanomy}) \cite{carlson2015stochastic}:
\begin{equation}\label{eq:SSD:sign}
x^{k+1} = x^k - \gamma_k \|d^k\|_1 \sign(d^k).
\end{equation}
The update is dense as each nonzero coordinate of $d^k$ produces the same magnitude change in $x^{k+1}$.

The matrix analog of the $\ell_\infty$-norm is the Schatten-$\infty$ matrix norm (also known as the spectral norm) which, if chosen to induce $h_k$, results in the stochastic spectral descent method \cite{carlson2015stochastic}:
\begin{equation*}
x^{k+1} = x^k - \gamma_k \|\sigma^k\|_1 U^k (V^k)^\top, 
\end{equation*}
where the reduced singular value decomposition (SVD) is given as $d^k = U^k \diag(\sigma^k) (V^k)^\top$.

For geometries arising from an inner product, which can represent both the updates of \ref{eq:SGD} and \ref{eq:PreSGD}, the sharp-operator is linear. In contrast,  the sharp-operator induced by a non-Euclidean geometry is notably nonlinear as explicitly illustrated in Table \ref{table:taxanomy} though the master template \eqref{eq:gen_form}. Nonetheless, preconditioning ideas can be naturally extended to this non-linear setting, as discussed in \cite{carlson2015preconditioned}.

We also note that is possible to construct a non-Euclidean norm tailored to the task of training deep neural networks by bounding the weight matrices layerwise in the spectral norm \cite{carlson2016stochastic}.
We will later revisit and motivate this idea in the context of feature learning.

\subsubsection*{Conditional Gradient Methods}

\looseness=-1While deterministic steepest descent is guaranteed to decrease the function value $f(x^k)$ at every iteration, it comes at the cost of picking a specific stepsize which is a function of the Lipschitz modulus $L$ of the gradient (specifically $\gamma_k < \nicefrac{2}{L}$).
In practice, this leads to slower training, since typically the global Lipschitz modulus is too pessimistic  where the function is less volatile.

\looseness=-1We can address this dependence on the Lipschitz constant by normalizing the gradient's magnitude and considers only its direction.
An example is the Normalized SGD (NSGD) with decoupled weight decay:
\begin{equation*}\label{eq:NSGD}
\tag{NSGD}
x^{k+1} = (1- \underbrace{\lambda_k }_{\color{blue} =s_k\lambda})  x^k - \underbrace{\gamma_k}_{\color{blue}=s_k\gamma } \tfrac{\nabla f(x^k)}{\|\nabla f(x^k)\|_2}, %
\end{equation*}
which uses the $\ell_2$-norm normalization. A very related technique in practice is known as \emph{gradient norm clipping}, which only normalizes the gradient when its $\ell_2$-norm exceeds a predefined threshold.

A natural generalization of \ref{eq:NSGD} to non-Euclidean norms follows from noticing that the normalization can be written in terms of a Linear Minimization Oracle ($\lmo$)
\begin{equation}\label{eq:lmo}\tag{$\lmo$}
\lmo_\rho(d) \in \argmin_{x \in \mathcal D} \braket{d,x}, 
\end{equation}
where the constraint is a (possibly non-Euclidean) norm-ball $\mathcal D := \{ x \mid \|x\| \leq \rho \}$. When $\rho=1$, we will omit it in the notation. 
Indeed, by choosing the $\ell_2$-norm ball with radius $\rho= \gamma/\lambda$ for $\mathcal{D}$, \ref{eq:NSGD} can be seen as an instance of the so-called Stochastic Conditional Gradient method (SCG) \cite{frank1956algorithm,ken-fw,jaggi2013revisiting,mokhtari2020stochastic,pethick2025trainingdeeplearningmodels},
\begin{equation*}\label{eq:SCG}
\tag{SCG}
\begin{split}
x^{k+1} &= (1-\lambda_k) x^k + \lambda_k\lmo_\rho(d^k)= (1-\lambda_k) x^k + \lambda_k\rho \lmo(d^k),
\end{split}
\end{equation*}
which requires $\lambda_k\in[0,1]$ and $x^0\in\mathcal{D}$ to ensure the iterates satisfy constraint $x^k\in\mathcal{D}.$ 
We refer to these class of methods as conditional gradient methods since they are \emph{conditioned} on the constraint.
This algorithm is also known as the stochastic Frank-Wolfe algorithm, where $\lambda_k$ is referred to as Frank-Wolfe stepsize. \ref{eq:SCG} can be used to provide numerical solutions to the constrained problem \eqref{eq:min} with $\mathcal{X}\equiv\mathcal{D}$.

\subsubsection*{A twist in perspective}  In addition to \ref{eq:NSGD}, the algorithms Lion, Signum, Muon, LARS and LAMB with decoupled weight decay can all be written as instances of the stochastic conditional gradient method (\ref{eq:SCG}) %
as shown in \cite{chen2023lion,pethick2025trainingdeeplearningmodels}. Intriguingly, \ref{eq:SCG} does not have a ``weight decay'' parameter. However, in light of their Frank--Wolfe interpretation, we can interpret that the roles of the stepsize and weight decay in these algorithms become inverted.  
The update direction corresponds to the output of the linear minimization oracle \ref{eq:lmo} over a ball of radius $\rho=\gamma/\lambda$ which does not depend on the dual feedback's magnitude.  
Consequently, the coefficient $\gamma$ in front of the normalized gradient no longer determines the actual progress along the optimization path as it merely sets the implicit radius of the feasible set.  
Instead, the contraction factor $(1-\lambda_k)$, commonly seen in the context of weight decay, \emph{and is interpreted as such}, plays the role of the \emph{Frank--Wolfe stepsize}, dictating how far the iterate moves toward the boundary of the norm ball at each step.  
Viewed through this lens, the so-called, decoupled ``weight decay'' parameter $\lambda$ is, geometrically, the true stepsize of the algorithm, whereas the nominal stepsize $\gamma$ only defines the scale of the constraint by scaling the $\lmo$.

\subsubsection*{Unconstrained variant} \looseness=-1
However, normalized gradient methods without weight decay also exist. Indeed, \ref{eq:NSGD} without weight decay can be seen as an instance of the unconstrained Stochastic Conditional Gradient method (uSCG) \cite{pethick2025trainingdeeplearningmodels}, which is also known as the continuous greedy method in submodular optimization: 
\begin{equation*}\label{eq:uSCG}
\tag{uSCG}
\begin{split}
x^{k+1} &= x^k + \gamma_k \lmo(d^k).
\end{split}
\end{equation*}

\begin{wrapfigure}{r}{0.40\textwidth}
    \vspace{-15pt}
    \centering
    \includegraphics[width=0.34\textwidth]{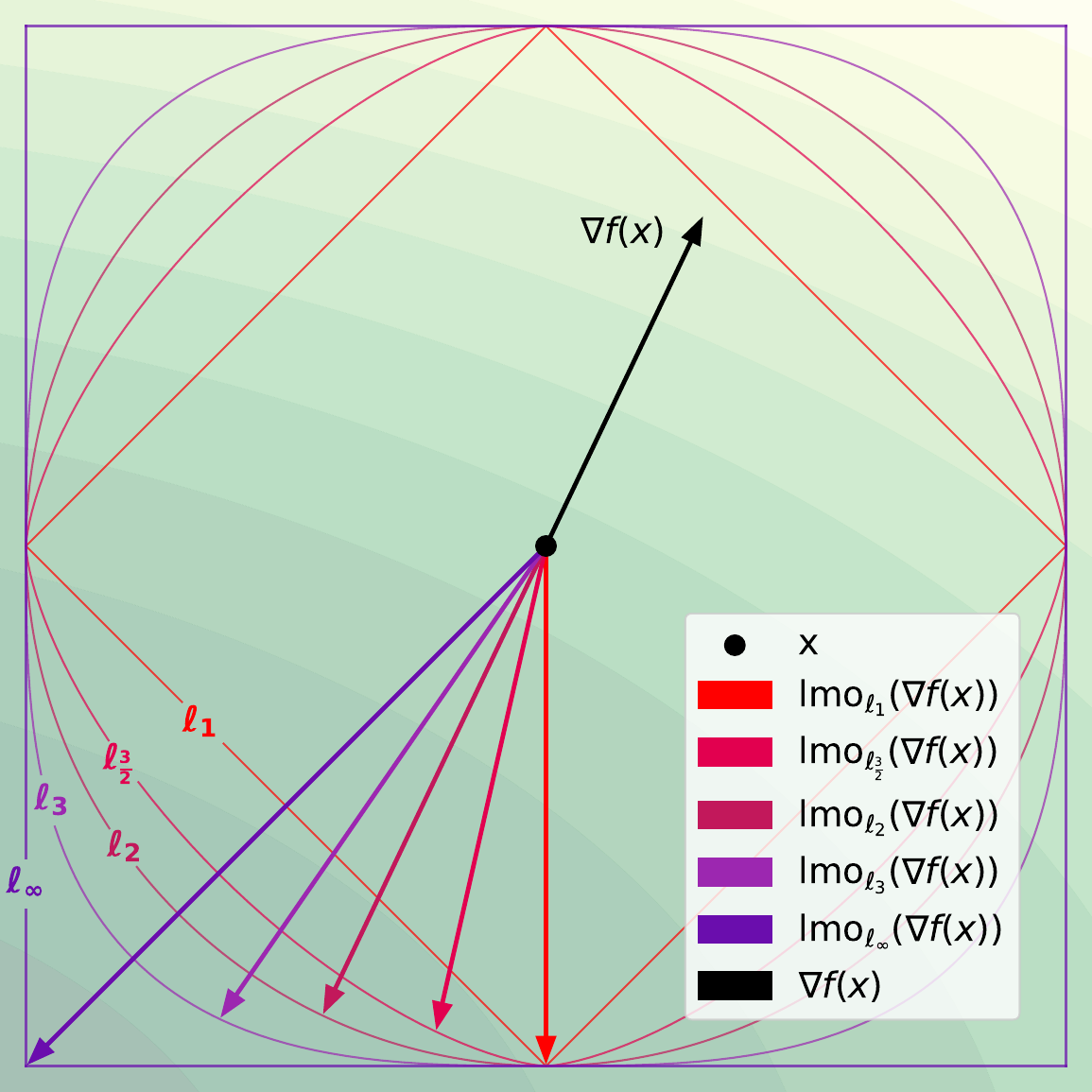}
    \caption{Visualization of the $\lmo$ for several different balls corresponding to $\ell_p$ norms for $p\in\{1,\tfrac{3}{2},2,3,+\infty\}$.}
    \label{fig:yourlabel}
    \vspace{-15pt}
\end{wrapfigure}

The \ref{eq:uSCG} method can also be seen as a trust-region method with a linear surrogate objective or as the generalized Matching Pursuit algorithm with a set of atoms $\mathcal{A}$ whose convex hull is $\mathcal{D}$, since the main operation used in those algorithms corresponds to the $\lmo$.
Whereas steepest descent is an instance of \eqref{eq:gen_form} with a regularizer $h_k$ that enforces a soft constraint, \ref{eq:uSCG} on the other hand is obtained by enforcing a hard constraint through the regularizer choice $h_k(x)=\iota_{\mathcal D}(x-x^k)$ where $\iota_{\mathcal D}(x):=0$ if $x \in \mathcal D$ and $+\infty$ otherwise.
This hard constraint explains the stability of the method, since the update never leaves the trust-region around the previous iterate $x^k$.

To see how \ref{eq:uSCG} can \emph{in general} be considered a normalized version of steepest descent, we point to the fact that the sharp operator and $\lmo$ can be defined in terms of each other.
Specifically, we have that
\begin{equation}\label{eq:lmo:sharp}
[d]^\sharp = -\|d\|_*\lmo(d).
\end{equation}
From this relationship we see that the $\lmo$ can be expressed as the normalized sharp-operator.

\paragraph*{Examples}
We can also deduce from the previous relationship that \ref{eq:uSCG} with $\ell_\infty$-norm constraints leads to the SignSGD update \cite{bernstein2018signsgd}: $x^{k+1} = x^k - \gamma_k \sign(d^k).$ This scheme was made popular in the distributed optimization literature due to the low communication cost of sending the bit-representable vector $\sign(d^k)$ as oppose to the floating point vector $d^k$. Similarly, for the spectral norm, \ref{eq:uSCG} reduces to $x^{k+1} = x^k - \gamma_k U^k(V^k)^\top.$
When $d^k$ is a momentum accumulation of past gradients, e.g., $d^k = (1-\alpha_k)d^{k-1} + \alpha_k g^k$, we recover the primary operation used in the Muon implementation \cite{jordan2024muon}.

\looseness=-1 From \eqref{eq:lmo:sharp} we also see a clear distinction between the $\lmo$ and the sharp operator. While the $\lmo$ is scale invariant (i.e., $\lmo(a\cdot s)=\lmo(s)$ for $a>0$), the sharp operator is scale equivariant (since $[a\cdot s]^\sharp=a[s]^\sharp$ for $a\in \R$).
The scale or affine invariance of conditional gradient methods makes the family of methods agnostic to the gradient size.
In particular, $\lmo$ outputs can always be chosen to lie on the boundary, i.e., $\|\lmo(s)\|=1$, such that the magnitude of the update in \ref{eq:uSCG} can be predetermined.

\begin{table*}[t]
\centering
\caption{Special instantiations of conditional gradient methods (\ref{eq:uSCG} and \ref{eq:SCG}) and \ref{eq:SSD} according to different choices of norm. 
The reduced SVD is given as $d=U\diag(\sigma) V^\top$ (here $d\in\mathbb{R}^{p_2\times p_1}$ is a matrix).}
\label{tbl:lmo}
\begin{tabular}{|l|l|c|c|c|c|}
\hline
& Method & $\alpha_k$  & Norm & $-\lmo(d)$ or $[d]^\sharp$ & Reference \\
\hline\hline
\multirow{6}{*}{\STAB{\rotatebox[origin=c]{90}{\scriptsize \hspace{10pt} Conditional Gradient}}}
  & Normalized SGD & $1$  & Euclidean $\|\cdot\|_2$ & $\tfrac{d}{\|d\|_2}$ & \cite{hazan2015beyond} \\
  & Normalized SGD with momentum & $[0,1]$  & Euclidean $\|\cdot\|_2$ & $\tfrac{d}{\|d\|_2}$ & \cite{cutkosky2020momentum}\\
  \cline{2-6}
  & SignSGD & $1$  & Max-norm $\|\cdot\|_\infty$ & $\sign(d)$ & \cite{bernstein2018signsgd} \\
  & Signum & $[0,1]$  & Max-norm $\|\cdot\|_\infty$ & $\sign(d)$ & \cite{bernstein2018signsgd} \\
  \cline{2-6}
  & Scion, Muon$\blue{^1}$ & $[0,1]$ & Layerwise Spectral $\|\cdot\|_{\mathcal{S}_\infty}$ & $UV^\top$ & \cite{pethick2025trainingdeeplearningmodels,jordan2024muon} \\
\hline\hline
\multirow{7}{*}{\STAB{\rotatebox[origin=c]{90}{\scriptsize \hspace{10pt} Steepest descent}}}
  & SGD & $1$  & Euclidean $\|\cdot\|_2$ & $d$ & \cite{Polyak1987} \\
  & SGD with momentum & $[0,1]$  & Euclidean $\|\cdot\|_2$ & $d$ & \\
  \cline{2-6}
  & \ref{eq:PreSGD} (AdaGrad, RMSProp) & $1$  & Mahalanobis $\|\cdot\|_{2,H}$ & $(H)^{-1/2}d$ & \cite{duchi2011adaptive,mcmahan2010adaptive,hinton2012neural} \\
  & \ref{eq:PreSGD} with momentum (Adam/AdamW) & $[0,1]$  & Mahalanobis $\|\cdot\|_{2,H}$ & $(H)^{-1/2}d$ & \cite{kingma2014adam,loshchilov2017decoupled} \\
  \cline{2-6}
  & Norm-scaled SignSGD & $1$  & Max-norm $\|\cdot\|_\infty$ & $\|d\|_1\sign(d)$ & \cite{kelner2014almost} \\
  & Norm-scaled SignSGD with momentum & $[0,1]$  & Max-norm $\|\cdot\|_\infty$ & $\|d\|_1\sign(d)$ &  \\
  \cline{2-6}
  & Stoch. Spectral descent & $1$ & Layerwise Spectral $\|\cdot\|_{\mathcal{S}_\infty}$ & $\|\sigma\|_1UV^\top$ & \cite{carlson2016stochastic} \\
  & Stoch. Spectral descent with momentum & $[0,1]$ & Layerwise Spectral $\|\cdot\|_{\mathcal{S}_\infty}$ & $\|\sigma\|_1UV^\top$ & \cite{lau2025polargrad} \\
\hline
\end{tabular}
\\
\footnotesize 
$\blue{^1}$ With non-Nesterov based momentum and ignoring the prescribed use of AdamW for some parameters. 
\end{table*}

\begin{figure}
    \centering
    \includegraphics[width=0.5\linewidth]{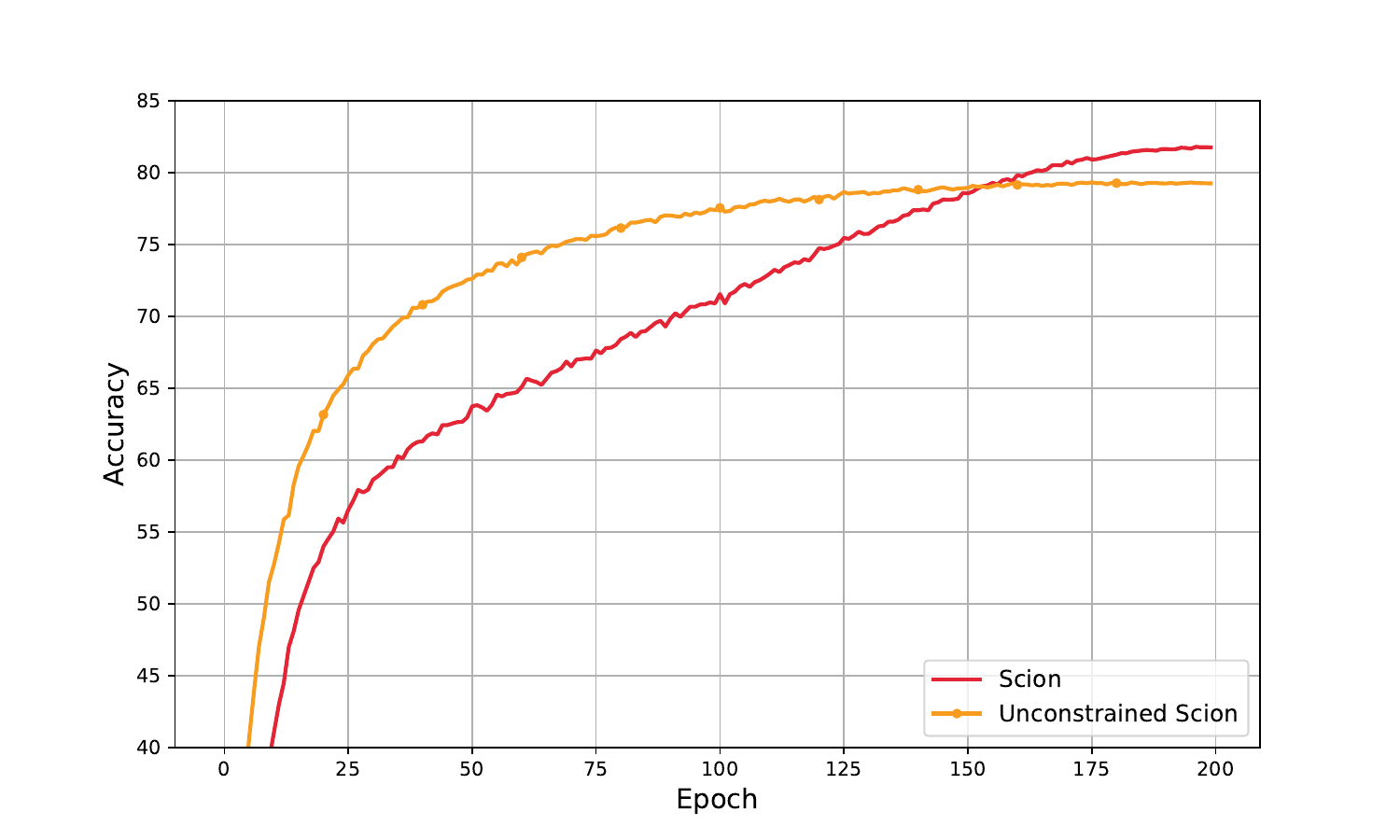}%
    \includegraphics[width=0.45\linewidth]{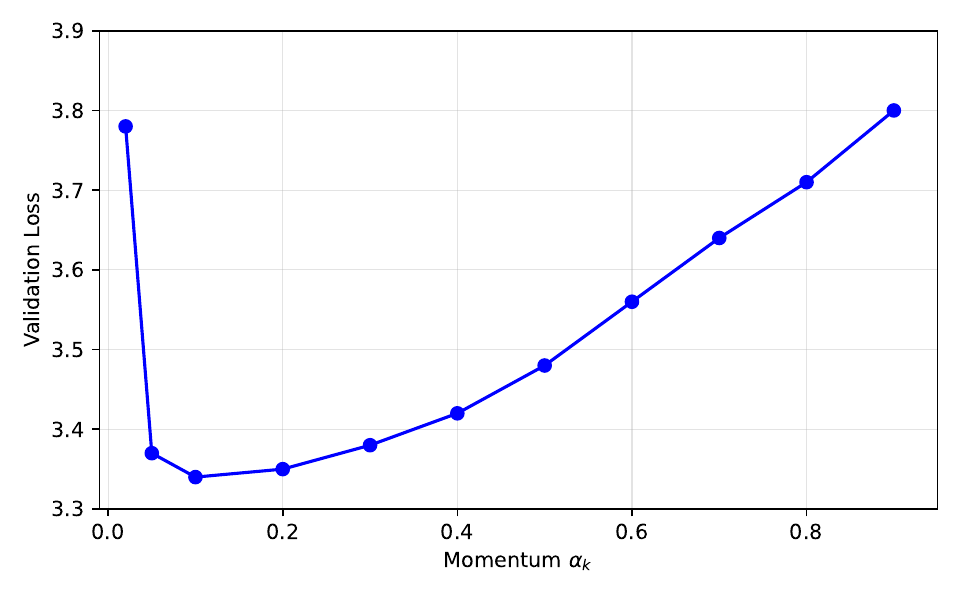}
    \caption{(left) The validation accuracy of a Vision Transformer on ImageNet benefits substantially from the tighter norm control provided by an instance of SCG (Scion) compared with uSCG (Unconstrained Scion) \cite{pethick2025trainingdeeplearningmodels}.
    (right) Momentum sweep on NanoGPT, where the stepsize $\gamma$ is retuned for each configuration. Performance reduces as momentum is disabled ($\alpha \rightarrow 0$).}
    \label{fig:wd+momentum}
\end{figure}

\tp{
\subsubsection*{Norm control of Conditional Gradient Methods}
Both \ref{eq:uSCG} and \ref{eq:SCG} provide a control of the norm of the iterates through $\|x^k\| \leq \rho\sum_{i=1}^k \gamma_i$ and $\|x^k\| \leq \rho$ for all $k\in \mathbb N$, respectively.
Specifically, \ref{eq:SCG} solves the \emph{constrained} problem %
\begin{equation}\label{eq:min:constrained}
\min_{x\in\mathcal{D}_{\rho}} f(x).
\end{equation}
Importantly, the control on the iterates is in the \emph{chosen norm}.
This norm control acts as a regularizer which turns out to be superior to placing an (Euclidean) $\ell_2$-regularization on the objective itself, which has led to the popularity of the AdamW optimizer, which replaces the $\ell_2$-regularization present in the Adam optimizer with decoupled weight decay.

In practice, controlling the norm is not only important for avoiding overfitting but also for providing numerical stability, especially for long training runs or under low-precision.
Long training runs are particularly common for generative adversarial networks (GANs), diffusion models, and in the pre-training of large language models.
We illustrate the importance of the norm control in \Cref{fig:wd+momentum}.
A similar benefit of decoupled weight decay can be observed for spectral methods as shown in \cite{pethick2025trainingdeeplearningmodels,liu2025muon}, which was later used to train the 1T parameter Kimi-K2 model through Muon with decoupled weight decay.
}

\subsubsection*{Frank-Wolfe vs.\ PyTorch weight decay}
\looseness=-1In \ref{eq:SCG}, the update employs three distinct scalars: a learning-rate schedule multiplier $s_k$, a Frank-Wolfe stepsize $\lambda$ (with $\lambda_k= s_k\lambda$), and a constraint radius $\rho = \gamma/\lambda$.  As we will see, under $\mu$-parameterization, we will choose scalings of the hyperparameters like the learning rate layerwise to zero-shot transfer them to larger models, which is known as $\mu$-transfer \cite{yang2022tensor}.  In the Frank-Wolfe view, the stepsize $\lambda$ is treated separately from the base radius $\rho$ (and schedule $s_k$), which will enable a clean hyperparameter transfer. 
PyTorch’s default implementation of weight decay instead folds these into each other, and changing \texttt{lr} for $\mu$-transfer unintentionally changes the effective decay: 
\begin{equation*}\label{eq:SCGW}
\tag{SCG-PyTorch}
\begin{split}
x^{k+1} &= (1-\texttt{lr}\cdot\texttt{weight\_decay}) x^k + \texttt{lr} \lmo(d^k).
\end{split}
\end{equation*}
The fix is trivial: treat $1/\texttt{weight\_decay}$ as the constraint radius $\rho$ and enforce \(\texttt{lr}_k = s_k \lambda /\texttt{weight\_decay}  \). 
With this, PyTorch and the Frank--Wolfe formulation are equivalent and both $\mu$-transfer correctly.

For any positive $\texttt{weight\_decay}$, it is similarly possible to interpret many other normalized algorithmic updates with PyTorch weight decay added within the \ref{eq:SCG} framework (\textit{cf.}, Table \ref{tbl:lmo}). In particular,  adding the weight decay simply modifies the constraint by scaling the radius to get an effective constraint set $\mathcal{D}_{1/\texttt{weight\_decay}}$ whose radius is $\tfrac{1}{\texttt{weight\_decay}}$ and a modified stepsize $\texttt{lr}\cdot\texttt{weight\_decay}$.

\subsubsection*{Momentum and Dual Averaging}
Momentum, as we have defined by
\begin{equation}\label{eq:momentum}
d^k = (1-\alpha_k)d^{k-1} + \alpha_k \nabla f(x^k,\xi_k),
\end{equation}
has a clear motivation in the non-Euclidean setting \cite{mokhtari2020stochastic,cutkosky2020momentum,pethick2025trainingdeeplearningmodels} since both the $\lmo$ and the sharp-operator may be biased. This can happen even if the SFO, $\nabla f(x^k,\xi_k)$ is unbiased, because of the nonlinearity of the $\lmo$ or the sharp-operator.
With careful choice of vanishing $\alpha_k$, the momentum estimator %
acts as variance reduction and can be used to guarantee convergence of \ref{eq:uSCG} \cite{pethick2025trainingdeeplearningmodels}.
In this sense, momentum provides a practical alternative to increasing batch sizes, which may not otherwise be practical.

The empirical benefits of momentum are evidenced by its presence in countless popular optimizers for deep learning.
This use of momentum is also well-motivated theoretically.
For the generic template \ref{eq:SCG} and \ref{eq:uSCG} with a constant stepsize $\gamma=\Theta(\frac{1}{n^{3/4}})$ and vanishing momentum $\alpha_{k}=\tfrac{1}{\sqrt{k}}$ it is possible to show convergence to a first-order stationary point with the rate
\begin{equation}\label{eq:uSCG:rate}
O\left(\tfrac{1}{n^{1/4}} + \tfrac{L}{n^{3/4}}\right),
\end{equation}
for $\bar x^k$ sampled uniformly over the iterates $\{x^k\}_{k\in [n]}$ \cite{pethick2025trainingdeeplearningmodels}.
Without momentum ($\alpha_k=1$) the algorithm would only be guaranteed to converge to a noise-dominated region \cite{pethick2025trainingdeeplearningmodels}.

\tp{
What does the rate in \eqref{eq:uSCG:rate} tell us? 
Since the $O(n^{-1/4})$ term asymptotically dominates, to see the benefit of a favorably Lipschitz constant $L$, we need to be in a regime where either the noise is small enough (i.e., by taking large batch sizes) or where the horizon $n$ is small.
This aligns well with practice where methods exploiting non-Euclidean geometries, such as SignSGD, Adam, Muon and Scion, typically benefit from the regime of large batch size and short runs.
}

{
As mentioned, the momentum estimator plays a central role for establishing convergence by reducing the variance.
This leads one to wonder if we can design more sophisticated estimators?
The space of possible dual feedback mechanism is indeed rich, which we illustrate with two prominent examples:
\begin{equation*}
\begin{split}
\text{Lion}: &\ \begin{cases}
\tilde{d}^k = (1-\alpha_k)\tilde{d}^{k-1} + \alpha_k \nabla f(x^k,\xi_k),  & \\
d^k = (1-\beta_k)\tilde{d}^{k-1} + \beta_k \nabla f(x^k,\xi_k).
\end{cases} \\
\text{STORM}: &\ d^k = (1-\alpha_k)d^{k-1} + \alpha_k \nabla f(x^k, \xi_k) + (1-\alpha_k)(\nabla f(x^k, \xi_k) - \nabla f(x^{k-1}, \xi_k)).
\end{split}
\end{equation*}
The estimator in the Lion optimizer was discovered empirically through a symbolic search over algorithms.
Notably, it is similar to the estimator in optimistic follow the regularized leader in online learning which uses $\nabla f(x^k,\xi_k)+\sum_{i=1}^{k}\nabla f(x^i,\xi_i)$.
Lion with $\alpha_k=\beta_k$ is also commonly referred to as Nesterov momentum and is used, e.g., in Muon.
Empirically, the Lion estimator has been observed to perform better than the momentum estimator in \eqref{eq:momentum} (recovered with $\beta_k=0$), while theoretically its mechanism remains to be explained.

On the other hand, the STORM estimator was originally discovered theoretically as a variance reduction mechanism for accelerating the convergence rate on nonconvex problems by improving the classical $ O(n^{-1/4})$ rate e.g., appearing in \eqref{eq:uSCG:rate} to $O(n^{-1/3})$.
It was later used in the MARS optimizer, which used a stale version of STORM in order to avoid computing two gradients per iteration (i.e., $\nabla f(x^{k-1},\xi_k)$ was replaced by the previously computed $\nabla f(x^{k-1}, \xi_{k-1})$).
Further improvements could be expected in finite-sum problems (i.e., the multi-epochs setting) by for example integrating the SPIDER estimator as in SPIDER-FW \cite{yurtsever2019conditional}.

Two themes in the design of dual feedback are averaging gradients and averaging iterates, which are equivalent in the Euclidean setting, since the $\lmo$ is then linear.
In the non-Euclidean setting, the nonlinearity of the $\lmo$ means the two are no longer equivalent, and the role of each becomes more clear.
As we shall see next, averaging the \emph{iterates} is important in producing the final model that is used.
}

\subsubsection*{Selecting a candidate solution}
An attentive reader might have noticed that the guarantee in \eqref{eq:uSCG:rate} for the stochastic case holds only for the average over the trajectory.
There are roughly two paradigms for providing guarantees for a final, candidate iterate: either ($i$) a new sequence of iterates is produced (e.g., by averaging), or ($ii$) the stepsize schedule is modified.
{Maybe surprisingly, the latter approach can also be viewed as a different way of averaging the iterates.}

A classical approach under convexity is to output the Polyak-Ruppert average, $\hat{x}^n:=\tfrac{1}{n}\sum_{k=1}^n x^k$, which is justified by Jensen's inequality.
In the non-convex case, this is unfortunately no longer justified theoretically, although variants of this are still used.
Similar to RMSProp, it is common in deep learning to instead output an exponentially moving average (EMA), $\hat{x}^{k+1}=\tau \hat{x}^{k+1} + (1-\tau) x^k$ where typically $\tau \in \{0.999, 0.9999\}$.

An alternative approach, when the horizon is known, is through the choice of the stepsize schedule.
Popular choices includes the cosine decay and the linear decay, both of which importantly decay the stepsize to zero (i.e., $\gamma_n=0$).
For the linear decay $\gamma_k = \gamma(1-k/n)$, the schedule is well-motivated in the convex setting, as it provably leads to last iterate guarantees for both non-smooth and stochastic problems \cite{zamani2023exact,defazio2024road}.
A practical advantage of this scheduling approach is that no additional memory is required, as would otherwise be the case when maintaining an averaged iterate like $\hat{x}^{k+1}$.

{
Remarkably, we can generalize and interpolate between these two seemingly distinct approaches \cite{defazio2024road}.
The generalization is obtained through the online learning framework and leads to the so-called \emph{schedule-free} method, which in addition to the iterates $x^{k}$ themselves maintains an average over the iterates $z^{k}$:
\begin{equation*}
y^k = (1-\beta_k) x^k + \beta_k z^k, \quad 
z^{k+1} = (1-c_{k+1})z^{k+1} + c_{k+1}x^{k+1}
\end{equation*}
where $c_{k+1} = \tfrac{1}{k+1}$ and $x^{k+1}$ is produced by some base optimizer receiving the feedback $\nabla f(y^k,\xi_k)$, e.g., \ref{eq:SCG}.
We recover the mentioned Polyak-Ruppert averaging with $\beta_k=0$, 
Polyak momentum with $\beta_k=1$,
and the linear decay stepsize schedule with $\beta_k=k/n$ where $n$ is the horizon.

As suggested by the dependency on $n$, linear decay only guarantees good performance for the \emph{final} iteration at the end of training, in contrast with a constant $\beta_k=\beta$, which provides an any-time guarantee.
Empirically, it has been observed that $\beta\approx 0.9$ works well, but the role of $\beta$ is still not fully understood. %

A reader familiar with optimization might recognize the schedule-free method is essentially a reparameterization of Nesterov momentum.
Interestingly, this perspective provides a connection to recent work on Lookahead-based optimizers, which similarly wraps a base optimizer, but executes multiple inner iterations before applying the outer iteration. DiLoCo uses Nesterov momentum for the outer loop and has been observed to improve performance significantly even in the single-client case, despite being developed for the distributed (multi-client) case.
}

\begin{wrapfigure}{r}{0.5\textwidth}
\centering
\vspace{-.5cm}
\includegraphics[width=0.48\textwidth]{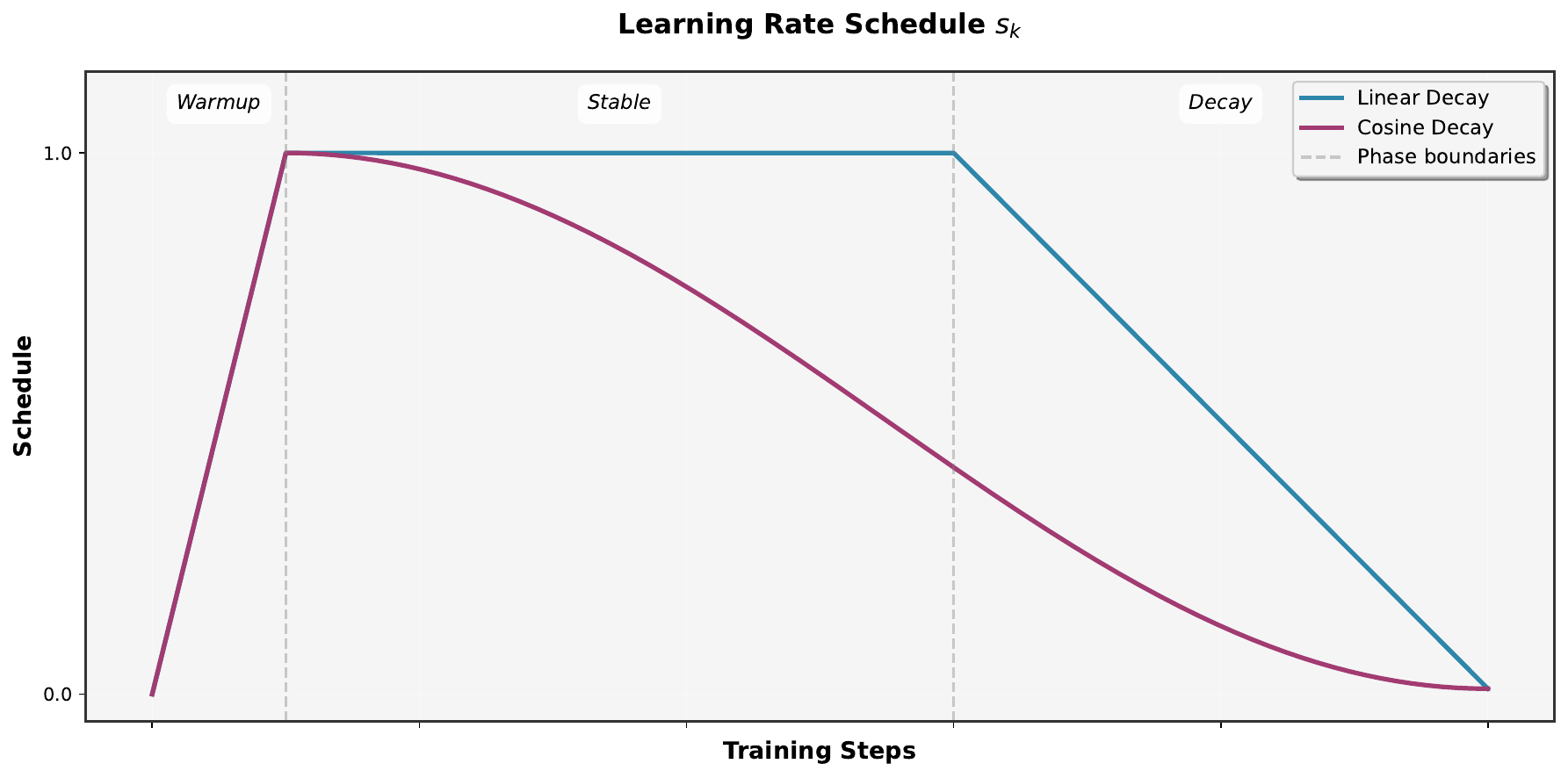}
\vspace{-0.5cm}
\caption{Typical learning rate schedules: linear and cosine decay.}
\vspace{-0.25cm}
\label{fig:lr_schedule}
\end{wrapfigure}

Apart from this \emph{cooldown} phase where the stepsize decays, there are typically one or two other phases in a scheduler.
First, it is common to have a \emph{stable} phase before cooldown in which the stepsize is taken constant.
Before this stable phase, an initial linear warmup of the stepsize for SGD and Adam is also common. %
Interestingly, the warmup phase does not seem to be necessary for optimizers with a priori adaptation to a favorable geometry, nor for long runs when using Adam. These common schedules are visualized in \cref{fig:lr_schedule}.%

\subsubsection*{Computational complexity}
\looseness=1 The computational cost of steepest descent and conditional gradient methods depends on the specific norm that induces the $\lmo$.
The sharp-operator used in steepest descent is only as expensive as the $\lmo$, since the additional dual norm scaling can be computed as $-\|d\|_* = \braket{d,\lmo(d)}=\operatorname{vec}(d)^\top \operatorname{vec}(\lmo(d))$ due to the definition of the dual norm and the optimality of $\lmo(d)$.
When $\|\cdot\|$ is a vector norm, the computation of the $\lmo$ is typically free, as is the case for $\ell_\infty$-norm (SignSGD) and $\ell_2$-norm (Normalized SGD).

{
However, the cost of the $\lmo$ can be significant when considering matrix norms.
In particular, let us focus on the $\lmo$ for the spectral norm, which is used in many modern optimizers (Table~\ref{tbl:lmo}):
$$U\diag(\sigma)V^\top \mapsto UV^\top\in \R^{p_2\times p_1}$$
One way of computing the spectral $\lmo$ is through the (reduced) singular value decomposition (SVD), which in the worse case has a slow $O(p_2p_1\cdot\min(p_2,p_1))$ time complexity.
Efficient alternatives to computing $UV^\top$ have fortunately been studied for decades.
We highlight three prominent families:

\begin{itemize}
\item \emph{Matrix-multiplication based} routines such as Newton-Schultz and PolarExpress \cite{amsel2025polar}.
One major benefit is that they are GPU friendly and work in low-precision.
Newton-Schultz is the solver used in Muon, where the the coefficients of the solver were hand-tuned to find a coarse approximation fast. PolarExpress further optimized these coefficients analytically.
\item \emph{QR based} routines such as power iteration used in the Dion optimizer \cite{ahn2025dion}, QDWH and ZOLO-PD, which have favorable properties in distributed settings, since the QR decomposition can be parallelized across devices.
One downside is that the QR decomposition requires full precision.
\item \emph{Sketching}, which is a randomized method that constructs a low-rank approximation to the SVD. 
This was the solver used originally in the stochastic spectral descent method (SSD) \cite{carlson2016stochastic}. %
\end{itemize}
Sketching and QR based methods have the additional benefit of enabling variants with efficient communication between devices by being able to truncate $U$ and $V^\top$, since they are materialized.
Another difference with the other methods is that, if the input is low-rank, then the output remains low-rank, which is in contrast with Newton-Schultz and PolarExpress for which the rank may increase to full.
}

\section*{Scaling Deep Learning Models}\label{sec:norm}

In the previous section we have covered how optimization methods can be designed to exploit the geometry of the optimization problem in general. In this section, we focus on neural networks and discuss how one can leverage theoretical tools to identify effective strategies to scale up deep learning models. We will see that such scaling rules often yield desirable practical benefits such as monotonic improvement in performance with scale and the transferrability of certain hyperparameters such as learning rate from small to large scale models \cite{yang2022tensor}. 
We will also see how considering the propagation of signals through the neural networks guides our choice of norm for optimizing the network.

\subsection*{Background}

Deep learning models are increasingly deployed at large scales. This is currently driven by the emergence of empirical neural scaling laws, which show that model performance predictably improves with scale \cite{hoffmann2022training} 
and follows a simple power-law relationship with the amount of available compute. These scaling laws play a crucial role in the design of foundation models in practice.

\begin{wrapfigure}{r}{0.39\textwidth}
    \centering
    \includegraphics[width=0.35\textwidth]{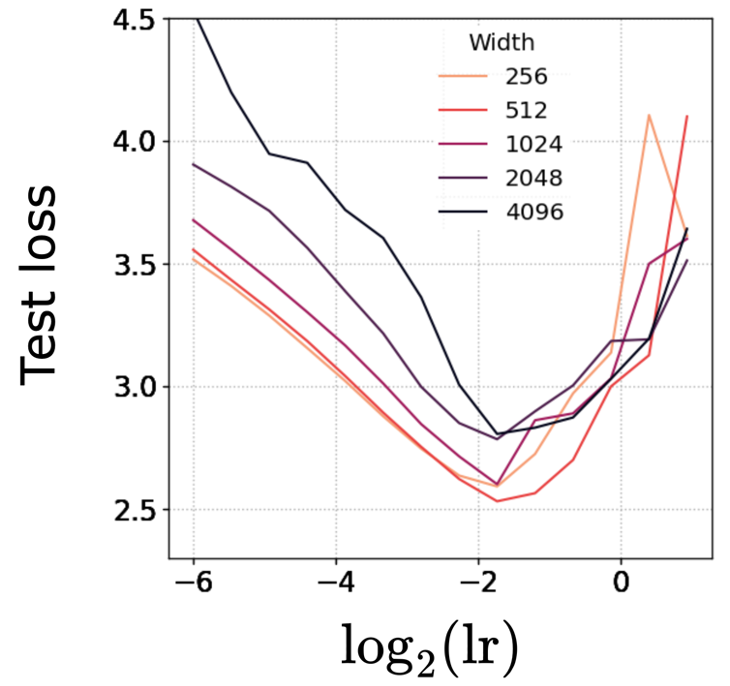}
    \vspace{-1em}
    \caption{Figure from \cite{vankadara2024feature}. Test loss against learning rate on Mamba with varying widths under \textit{standard scaling}. Performance of the optimally tuned model starts to worsen after a certain width.} 
    \label{fig:mamba-worsens-width}
    \vspace{-1em}
\end{wrapfigure}

\looseness=-1 Since training at such scales typically requires substantial financial and computational resources for even a single run, design choices are primarily informed by extrapolating from smaller models, allowing researchers to compare alternative design choices—such as Adam versus Shampoo—without exhaustive large-scale experiments. Beyond improving performance, increasing scale also appears to induce qualitatively novely behaviors, known as emergent capabilities, suggesting that scaling is not merely a quantitative phenomenon \cite{wei2022emergent}. 

Crucially, however, scaling does not always yield better performing models, and in general arbitrary scaling can hurt model performance. For instance, \cite{vankadara2024feature} shows that if we increase the width of a Mamba model \cite{gu2023mamba} according to the \textit{default scaling rule} used in practice, performance of the model deteriorates after a certain width threshold (see Figure \ref{fig:mamba-worsens-width} for an illustration). Examples such as these are plentiful in literature \cite{yang2022tensor}.

\begin{center}
    To ensure improved model performance with scale, we need  \textit{principled scaling rules}. 
\end{center}

As the available computational budget increases, the training configuration of a neural network can be adapted in countless ways. These include scaling dimensions such as network width, depth, dataset size, or batch size. A scaling rule prescribes how key aspects of training—such as the architecture, hyperparameters, and dataset—should be adjusted as the computational budget varies.

To scale neural networks effectively, it is essential to establish scaling rules that satisfy at least two key desiderata: 
\begin{enumerate}
    \item \textit{Optimality:} The large-scale model should be highly performant relative to alternative scaling rules.
    \item \textit{Predictability:} Performance of large-scale models can be reliably inferred from small-scale models,  facilitating efficient exploration of the design space.
\end{enumerate}

 \textit{Scaling limits} serve as a powerful theoretical framework to derive such scaling rules and more generally to understand the behaviour of large scale models. Scaling limits are the limits of neural networks obtained by taking various scaling dimensions such as the width or depth of the network to infinity.
 
  In principle, to achieve optimality, one can derive the scaling limits associated with different scaling rules and determine which ones yield the most favorable asymptotic properties, such as stability of the training dynamics and feature learning. Predictability, on the other hand, requires that models rapidly converge to their scaling limits, motivating an analysis of the rate of convergence under various scaling rules. 
  
  Given the complexity of deriving scaling limits for arbitrary scaling strategies, analyses often constrain the space of scaling rules under consideration. In this section, we primarily focus on scaling rules where only the width of the network is taken to infinity, through there is growing literature covering the increasing depth as well \cite{dey2025don,large2024scalable}.

  Developing theoretical tools to analyze these scaling limits has been an ongoing challenge in deep learning theory. One influential approach in this direction is the Neural Tangent Kernel (NTK) framework \cite{jacot2018neural}, which allows us to study the behaviour of infinitely wide neural networks under a specific choice of width scaling, known as the Neural Tangent Kernel Parameterization.
NTK theory has provided fundamental insights into deep learning dynamics. 
However, a key limitation of NTK analysis is its inability to capture \textit{feature learning}. In this regime, the parameters remain close to their initialization, and training dynamics reduce to kernel regression with a fixed, data-independent kernel. Consequently, the underlying feature representation is determined solely by the architecture and initialization rather than learned from data, which restricts the relevance of NTK theory to practical deep learning settings.

Early attempts to circumvent this limitation focused on mean-field theory, particularly for two-layer networks \cite{chizat}. %
More recently, the Maximal Update Parameterization ($\mu$P) has emerged as a principled generalization to deep architectures. 
Introduced by \cite{yang2021tensor}, $\mu$P is based on an expanded space of hyperparameters that allows for different learning rates and initialization variances in each layer of the neural network.
This parameterization enables nontrivial feature evolution even in the infinite-width limit, thus bridging the gap between theory and practical deep learning.

The foundation of $\mu$P is the Tensor Programs framework \cite{yang2021tensor}, which provides a systematic way to analyze the asymptotic behavior of wide neural networks at both initialization and during training. 
Tensor Programs are defined through a structured sequence of operations, including matrix multiplications, non-linear outer products, and vector averages. 
This framework allows for a precise tracking of neural network dynamics, leading to the discovery of $\mu$P as the unique scaling rule that guarantees feature learning for every layer, regardless of width.
Furthermore, empirically, $\mu$P has been shown to enable the transferability of optimal hyperparameters like learning rate from small to large-scale models, a property that is highly desirable in large-scale deep learning applications.

A complementary perspective on feature learning is rooted in operator norms.
The authors of \cite{yang2023spectral} show that effective feature learning is achieved by scaling the spectral norm of weight matrices and their updates proportionally to the square root of the ratio between output dimension and input dimension.
The spectral condition recovers $\mu$P and, importantly, motivates the non-Euclidean optimization methods covered in the previous section.

\paragraph*{Outline}
This section first introduces the abc-parameterization and identifies the Maximal Update Parameterization ($\mu$P).
We then provide an intuitive understanding of $\mu$P through the spectral perspective by focusing on the infinite-width limit.
From this perspective we will clearly see the implication for initialization, norm constraint and the importance of the low-rank structure of gradients.
We close by discussing practical aspects of hyperparameter transfer, the connection with Lipschitz continuity, and the norm choice for the optimizer.

\subsection*{Identifying optimal scaling limits}
Consider the setting of an $L$-hidden-layer multilayer perceptron (MLP) defined iteratively via \[
h_1(z):= W_1 z, \qquad o_\ell(z):= \Phi(h_\ell(z)),\qquad h_{\ell+1}(z):= W_{\ell+1} o_\ell(z),\qquad \Phi(z):= W_{L+1} o_L(z),
\] for inputs $z\in \mathbb{R}^{p_{1}}$
with weight matrices $W_1\in\mathbb{R}^{p \times p_{1}}$, $W_\ell\in \mathbb{R}^{p \times p}$ for $\ell\in[2,L]$, and $W_{L+1}\in \mathbb{R}^{p_{L+1}\times p}$.
Given a loss function $\mathcal{L}(\Phi(z),y)$, we use the SGD update rule \eqref{eq:SGD} to update the parameters of the MLP
\begin{equation}
\tag{\sc SGD}
    \label{eq:SGD}
    W_\ell^{k+1} = W_\ell^k - \eta_\ell \nabla_{W_\ell^k} \mathcal{L}\left(\Phi(z),y\right). 
\end{equation}

To understand the scaling behaviour of MLPs trained under SGD, \cite{yang2021tensor} introduced a natural class of width-scaling rules---\textit{abc-parameterizations}---which allow layerwise scalings of the 
parameters, the variance of their initializations, and learning rates by introducing width-dependent scaling factors. Throughout, we will use $\Theta(\cdot)$ to represent the asymptotic order in the infinite width limit $p\to\infty$. 

\begin{definition}[\textbf{$abc$-parametrization} \cite{yang2021tensor}]\label{def:abc}
    An \textit{$abc$-parametrization} $\{a_\ell\}_{\ell\in[L+1]}\cup\{b_\ell\}_{\ell\in[L+1]}\cup\{c_\ell\}_{\ell\in[L+1]}$ parameterizes an MLP trained with SGD in the following way:
    \begin{enumerate}
    \item Parameterize the weight matrices $W_\ell$ of the MLP as $p^{-a_\ell} w_\ell$, for a trainable weight parameter $w_\ell$ of the same shape as $W_\ell$, for all $\ell \in [L+1]$.
        \item Initialize the trainable weights as $w_\ell\sim \mathcal{N}(0,p^{-2b_\ell} I)$.
        \item Train the weights using the SGD update rule with layerwise learning rates for some width independent $\eta$,
        \begin{align*}
    w_\ell^{k+1} = w_\ell^k - \eta p^{-c_\ell} \nabla_{w_\ell^k} \mathcal{L}\left(\Phi(z),y\right),
    \end{align*}
\end{enumerate}
\end{definition}

Most standard initialization schemes used in practice fall within this family of scaling rules. For instance, He/LeCun initializations are specified by $a_\ell = 0 \; \forall \ell \in [L+1], b_1 = 0 \text{ and } b_\ell = 1/2 \text{ for } \ell > 1, \text{ and } c_\ell = 0 \; \forall \ell \in [L+1]$, i.e., 
   \begin{align*}
        W_1 \sim \mathcal{N}(0, 1/p_1), \; 
        W_\ell \sim \mathcal{N}(0, 1/p),  \; \;  \text{for } \ell \in [2, L], \;
        W_{L+1} \sim \mathcal{N}(0, 1/p),
    \end{align*}
    with a global learning rate \(\eta_\ell = \eta\) for all \(\ell \in [L+1]\).
This parametrization, commonly referred to as the Standard Parametrization (SP), fails to yield width-independent feature evolution in the infinite-width limit when trained with Mean-Squared Error (MSE) loss.

To identify optimal scaling limits, one can analytically derive the infinite width limits of MLPs trained with SGD under this family of abc parameterizations and classify all possible limits according to their qualitative behaviour. To this end, the Tensor Program (TP) framework introduced by \cite{yang2021tensor} provides a way to derive the infinite-width limits of neural networks under arbitrary abc-parameterizations.

As the width of hidden layers approaches infinity, Tensor Programs identify two distinct dynamical regimes among others:
\begin{enumerate}[label=(\roman*)]
  \item \emph{Kernel Regime}: here, weight updates are small relative to initialization, and the network behaves like a kernel method where no features are learned. 
  The limiting behavior is governed by a fixed kernel derived from the initialization. Formally, an abc-parametrization \emph{is in the kernel regime} if there exists a positive semidefinite kernel $K$ such that, %
    in the $p\to\infty$ limit,
    \begin{equation}
        \Phi^{k+1}(z)=\Phi^{k}(z)-\eta K(z,z^{k})\mathcal{L}'({\Phi}^{k}(z^{k}),y^{k}),\quad \forall k\ge0.
        \label{eqn:kerneleqn}
    \end{equation}
  \item \emph{Feature Learning Regime}: weight updates are scaled to induce stable feature evolution, enabling the network to learn data-dependent representations. Formally, an abc-parametrization \emph{admits feature learning} if, as $p\to\infty$, each entry of $\Delta o_L(z)$ is $\Theta(1)$ where $\Delta o_L(z)$ denotes the change in the last layer activation vector $o_L(z)$ after one step of SGD. 
\end{enumerate}

\paragraph*{Maximal Update Parameterization (\(\mu\)P)}

Under the feature learning regime, there exists a unique \textit{abc}-parameterization (up to an equivalence class) that enables maximally stable feature learning. Formally, for all \(\ell \in [L]\), the update term \(\Delta W_\ell o_{\ell-1}(z)\) has \(\Theta(1)\) coordinates. In other words, weight updates in every layer induce a non-trivial effect on the network's output. As feature learning is considered an useful mechanism for generalization, one may argue that $\mu\text{P}$ is an optimal scaling rule within this class. 

For MLPs trained with SGD, \(\mu\)P can be specified by the parameter choices:  
\[
a_\ell = 0, \quad \forall \ell \in [L+1], \quad 
b_1 = 0, \quad b_\ell = \tfrac{1}{2}, \quad \text{for } \ell \in [2, L], \quad 
b_{L+1} = 1,
\]
\[
c_1 = -1, \quad c_\ell = 0, \quad \text{for } \ell \in [2, L], \quad 
c_{L+1} = 1.
\]
This corresponds to the following initialization and learning rate scalings:
\begin{equation}
  \begin{gathered}
  \text{Initialization:} \quad
        W_1\sim\mathcal{N}(0, {1}/{p_1}),\quad
        W_\ell \sim\mathcal{N}(0, {1}/{p}) \quad \text{for } \ell\in[2,L],\quad
        W_{L+1} \sim \mathcal{N}(0, {1}/{p^2}).
        \\
  \text{Layerwise SGD learning rates:} \quad
        \eta_{1} =  \eta p,\quad
        \eta_{\ell} = \eta,  \quad \text{for } \ell\in[2,L],\quad
        \eta_{{L+1}} = \eta p^{-1}.
  \end{gathered}
  \label{eq:MUPMLP}
\end{equation}

Alternative parameterizations that achieve maximally stable updates can be derived by leveraging the symmetries of \textit{abc}-parameterizations (see \cite{yang2021tensor}). Specifically, for any \(\alpha \in \mathbb{R}\), shifting the parameters as  
\[
a_\ell \to a_\ell + \alpha, \quad 
b_\ell \to b_\ell - \alpha, \quad 
c_\ell \to c_\ell - 2\alpha
\]
does not affect the training dynamics. That is, the activations and the function computed by the network remain unchanged.

 \(\mu\)P scaling offers several practical benefits. It often tends to improves performance and performance tends to monotonically improve with scale.  Empirically, it has been shown that, under \(\mu\)P, many properties of wide neural networks including loss curves, pointwise predictions, pre-activation distributions, internal representations etc tend to consistent across large widths
compared to other parameterizations, such as the Neural Tangent Kernel (NTK) parameterization, reflecting better predictability.
 
 Another particularly interesting practical benefit is that under \(\mu\)P, certain optimal hyperparameters such as learning rate can be transferred from small to large-scale models, which is far from standard practice. This also suggests that under \(\mu\)P, finite-sized models are potentially closer to their infinite-width limit.

\paragraph*{Deriving optimal scaling rules for other architectures and optimizers} 
\tp{While our discussion has evolved around MLPs, the framework of Tensor Programs can be utilized to analytically derive infinite width limits and accordingly identify optimal\footnote{It is not clear a priori what the notion of optimality entails (see for instance \cite{haas2024boldsymbolmumathbfp}). } scaling limits for a fairly broad class of architectures, such as ResNets and Transformer, and optimizers such as Adam.}
However, there are notable exceptions. Certain modern state space models, such as the Mamba architecture, cannot be represented as a Tensor Program. Rigorously establishing their infinite-width limits requires a significant generalization of the existing framework \cite{vankadara2024feature}.

\subsection*{Infinite-width limit \& the spectral condition}
\tp{Why are different layers treated differently in $\mu$P? To build intuition, we will see how feature learning can follow from a spectral conditions on the weights and updates in the special case of MLPs and normalized updates \cite{yang2023spectral}.
We will focus on the simplified case of linear MLPs, where the number of assumptions needed is minimal.}
We define the initial hidden layer by $h_1(z) = W_1z$ and the remaining layers by $h_\ell(z) = W_\ell h_{\ell-1}(z), \ \forall \ell \in 2,..,L+1;$. 
We denote the global loss as $\mathcal L(h_L(z),y)$ where $\mathcal L$ is the loss function and $y$ is a 1-hot encoded target vector.
The RMS norm, $\|z\|_\RMS:= \tfrac{1}{\sqrt{p}}\|z\|_2$ for $z\in \R^p$, will play a central role.
We can redefine feature learning in terms of the RMS norm:

\begin{definition}[Feature Learning]\label{def:feature}
Let $\Delta h_\ell(z)$ denote the change in the layer-$\ell$ activations.
Then we are said to be in the feature learning regime if the following holds for all $\ell \in [L+1]$:
\begin{enumerate}[label=(\roman*)]
  \item $\|h_\ell(z)\|_\RMS = \Theta(1)$ (the forward pass is stable)
  \item $\|\Delta h_\ell(z)\|_\RMS = \Theta(1)$ (the update is non-vanishing and non-exploding)
\end{enumerate}
\end{definition}
The definition states that the typical element of the vectors should be of order $\Theta(1)$ as $p$ goes to infinity.
If the entries were larger, then they would blow up with width, and thus become numerically unstable.
If they were any smaller this could likewise lead to instability.
Additionally, if $\|\Delta h_\ell(z)\|_\RMS$ is not lower bounded, and thus potentially vanishing with width, the model may be in the less expressive Neural Tangent Kernel (NTK) regime where features do not evolve.

What we control in practice are not the preactivations $h_\ell(z)$ themselves, but rather the weight matrices $W_\ell$ that induce them.
Fortunately, we can upper bound $\|h_\ell(z)\|_\RMS$ through a norm control on $W_\ell$.
Recall the operator norm defined as follows
\begin{equation}\label{eq:opnorm}
\|A\|_{\alpha \rightarrow \beta}
:=\max_{\omega \in \mathbb{R}^p, \omega \neq 0} \frac{\|A\omega\|_\beta}{\|\omega\|_\alpha}
= \sup_{\|\omega\|_\alpha = 1} \|A\omega\|_\beta.
\end{equation}
Directly from the definition, we see that if the input $\omega$ is bounded through $\|\omega\|_\alpha\leq 1$, then the output $\|A\omega\|_\beta$ will be bounded when $\|A\|_{\alpha \rightarrow \beta}$ is bounded.
Thus, $\|W_\ell h_{\ell-1}(z)\|_\RMS$ can be upper bounded through an upper bound on $\|W_\ell\|_{\RMS \rightarrow \RMS}$, which is just a rescaled version of the Schatten-$\infty$ norm (a.k.a. the spectral norm), i.e., $\|W_\ell\|_{\RMS \rightarrow \RMS}=\sqrt{\tfrac{p_{\mathrm{in}}}{p_{\mathrm{out}}}}\|W_\ell\|_{\mathcal{S}_\infty}$.
The only additional requirement is that the original input to the network is bounded, i.e., $\|z\|_\RMS < \infty$.
A similar reasoning applies for upper bounding $\|\Delta h_\ell(z)\|_\RMS$.%

If we additionally knew that these bounds were tight, then feature learning could be restated solely in terms of the weight matrices as follows:
\begin{definition}[Spectral condition]\label{def:spectral}
The spectral condition requires that 
\begin{enumerate}[label=(\roman*)]
\item $\|W_\ell\|_{\RMS \rightarrow \RMS}=\Theta(1)$ (\emph{the weight matrices are properly initialized})
\item $\|\Delta W_\ell\|_{\RMS \rightarrow \RMS}=\Theta(1)$  (\emph{the updates are properly scaled at each iteration})
\end{enumerate}
\end{definition}
\tp{
\paragraph*{Implication for initialization}
The spectral condition needs to hold for all the iterates throughout optimization.
At initialization, condition (i) can be trivially satisfied by directly initializing $W_\ell$ to be a scaled semi-orthonormal matrix or
through Gaussian initialization, $[W_\ell]_{ij} \sim \mathcal N(0,\sigma_\ell^2)$, since random
matrix theory tells us that the spectral norm scales as $\|W_\ell\|_{2\rightarrow 2} \approx \sigma_\ell (\sqrt{p_{\ell-1}} + \sqrt{p_{\ell}})$.
This motivates the following variance choice (\textit{cf}., \cite[Param. 1]{yang2023spectral}):
\begin{equation*}
\sigma^2 = \Theta(\tfrac{1}{p_{\ell-1}}\min \{1, \tfrac{p_{\ell}}{p_{\ell-1}}\}).
\end{equation*}
For language modeling, where the vocabulary size (i.e., the output dimension for the last layer) is much larger than the network width (i.e., input dimension for the last layer), the variance requirement above implies that even the last layer can be initialized with $\sigma^2 = \Theta(\tfrac{1}{p_{L}})$, which is satisfied by e.g., the popular Kaiming initialization \cite{he2015delving} otherwise typically associated with SP (see e.g., \Cref{fig:GPT}, which successfully uses SP initialization for hyperparameter transfer).

\paragraph*{Low-rank structure of gradients}
How do we show that the upper bounds provided through the operator norm are tight, thereby proving that the spectral condition implies feature learning?
The bounds can be particularly loose when the incoming vector $h_{\ell-1}(z)$ interacts with small singular values of the matrices $W_\ell$ and $\Delta W_\ell$.
Fortunately, $h_{\ell-1}(z)$ turns out to be aligned with the top singular vectors of the matrices.
The key insight is that the gradients happens to be rank-1 with right singular vector $h_{\ell-1}(z)$ when the batch size is one, so the alignment trivially holds.

Due to the rank-1 structure of the gradients, many update rules coincide for batch size one.
For instance, in this case the Frobenius norm and spectral norm of the gradients are equivalent, so the two corresponding $\lmo$-based updates are equivalent.
Furthermore, both are equivalent to normalizing with the spectral norm.
Thus, all three of these methods provide valid ways of ensuring that $\|\Delta W_\ell\|_{\RMS \rightarrow \RMS} = \Theta(1)$ in the single sample case.
}

For batch sizes greater than one, the equivalence between the methods no longer strictly holds and establishing alignment becomes more difficult.
However, despite not being rank-1, it has been observed that the stable rank remains small even for large batch sizes in practice.

\paragraph*{The importance of layerwise treatment}
One central takeaway from $\mu$P and the preceding reasoning in the infinite width limit is that weights should be optimized differently depending on which layer they correspond to in the network (e.g., first vs last).
Focusing on the spectral condition (\Cref{def:spectral}), which requires that $\|\Delta W_\ell\|_{\mathcal{S}_\infty}=\Theta(\sqrt{\tfrac{p_\ell}{p_{\ell-1}}})$, there are three distinct cases:
\begin{enumerate}[label=(\roman*)]
\item $\|\Delta W_1\|_{\mathcal{S}_\infty}=\Theta(\sqrt{p_1})$ (the input scaling grows)
\item $\|\Delta W_\ell\|_{\mathcal{S}_\infty}=\Theta(\sqrt{\tfrac{p_\ell}{p_{\ell-1}}})$ for all $\ell \in \{2,...,L+1\}$ (the hidden layers are constant with width)
\item $\|\Delta W_{L+1}\|_{\mathcal{S}_\infty}=\Theta(\tfrac{1}{\sqrt{p_{L}}})$ (the output scaling goes to zero)
\end{enumerate}
The distinction is a simple consequence of the input and output dimension being kept fixed, but it has nonetheless stark implications. 
For example, while the hidden layers should have constant magnitude updates, we see that the scaling of the last layer updates should instead approach zero as $p_{L}\rightarrow \infty$.

This reasoning through the spectral condition highlights many of the key ingredients, such as the choice of initialization and the importance of the low-rank structure of gradients.
\tp{However, the complete argument relies on a technical assumption stating that a certain cancellation of terms never occurs, requires additional assumptions to treat activation functions, and is restricted to MLPs and normalized updates. %
The Tensor Program framework provides a way to avoid these assumptions, %
and it is flexible enough to cover a large family of both networks and algorithms. %
}

\paragraph*{Verifying feature learning}

Suppose that we have parameterized our model and algorithm such that we should expect feature learning.
How do we verify in practice that our updates are indeed in the feature learning regime, and not instead unstable like SP or in the lazy regime?

\begin{wrapfigure}{r}{0.50\textwidth}
\centering
\includegraphics[width=\linewidth]{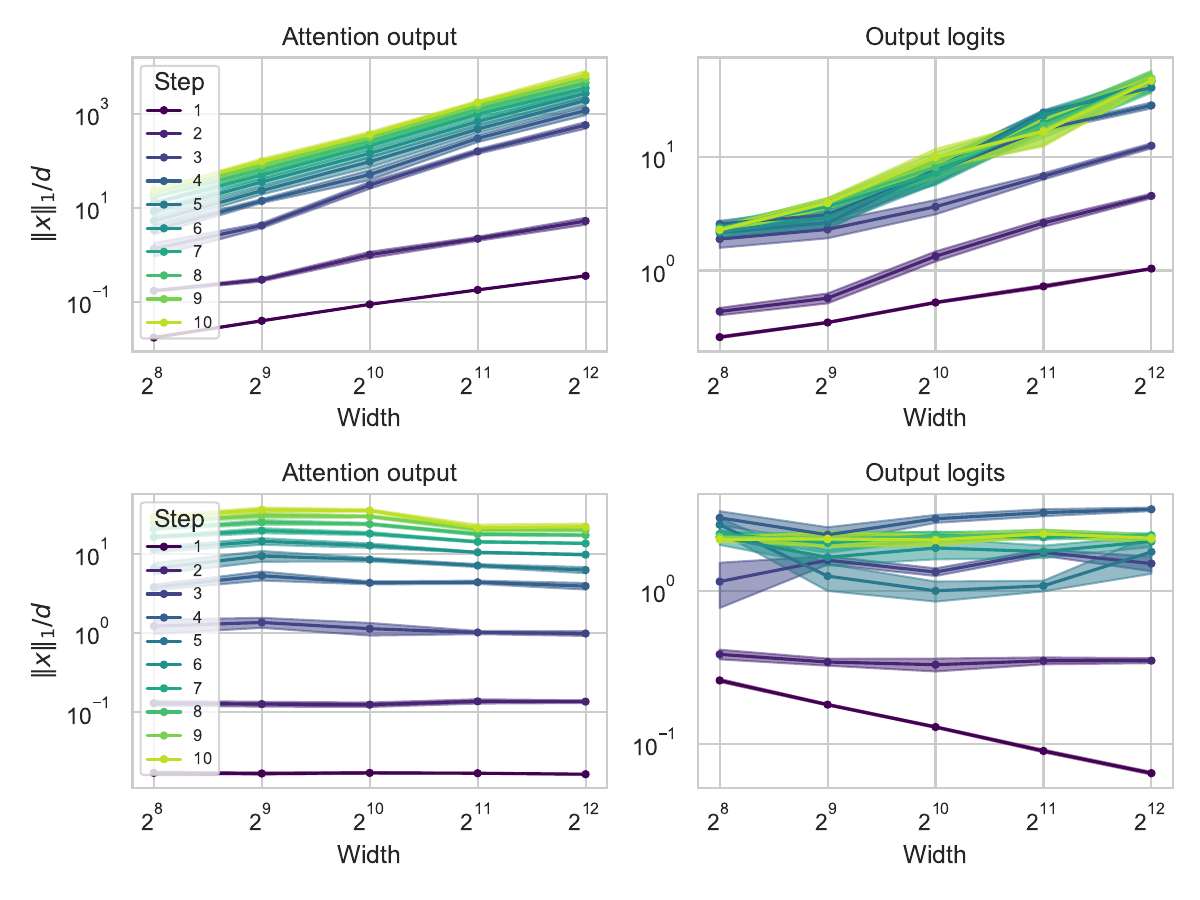}
\caption{Coordinate check on a 3-layer GPT model using the AdamW optimizer with SP (top) and $\mu$P (bottom). Preactivations and logits remains constant with width for $\mu$P as opposed to SP for which they blow up.}
\label{fig:coordcheck}
\end{wrapfigure}

The fact that hyperparameters transfer across model sizes can be an indication, but it can be misleading, since transfer might be observed without all layers exhibiting feature learning.
The appropriate test is instead the so-called \emph{coordinate check}, which is also significantly cheaper to perform.

In essence, a coordinate check simply verifies the feature learning definition, which states that the coordinates of preactivations throughout the network should remain constant with width.
This property can be verified by running a few steps of the algorithm across different model sizes, and either computing the $\ell_\RMS$-norm or the normalized $\ell_1$-norm of the preactivations, which measures the magnitude of the typical entry and the magnitude of the average entry, respectively.
In practice, these two metrics are very similar.
An example is provided in \Cref{fig:coordcheck} showing the clear distinction between Standard Parameterization (SP) and the Maximal Update Parameterization ($\mu$P).

\subsection*{Norm choice}

We have seen how feature learning can be ensured through the spectral condition. %
In this section, we cover how the requirements of the spectral condition connect to a norm choice on the full network parameters, and how that choice leads to an instantiation of the generic algorithmic framework introduced in the previous section.
\tp{Specifically, we will see that Lipschitz conditions can be shown for all layers of the neural network in a given norm, when the weights are bounded in the same norm.
The associated Lipschitz constant can importantly be made independent of the width, which provides a non-asymptotic argument for hyperparameter transfer.}
Finally, we touch on some of the limitations with this approach, since these Lipschitz conditions (upper bounds) are themselves not sufficient to rule out the lazy regime.

\paragraph*{A norm on the entire network}
The spectral condition (\Cref{def:spectral}) requires that the $\mathrm{RMS}\to\mathrm{RMS}$ norm of each layer's weight matrix is controlled and the same for each layer's update. The upper bound of these requirements can be satisfied simultaneously by choosing the max-norm on the joint parameters
$x=\{W_\ell\}_{\ell \in [L+1]}$:
\begin{equation}\label{eq:maxnorm}
\|x\|_\mathcal{X} = \max_{\ell \in [L+1]} \tfrac{1}{\rho_\ell}\|W_\ell\|_{\mathcal{W}_{\ell}}
\end{equation}
where each norm $\|\cdot\|_{\mathcal{W}_{\ell}}$ is a layer-dependent operator norm, e.g., $\mathrm{RMS}\to\mathrm{RMS}$.
With this choice, any algorithm that guarantees $\|x\|_{\mathcal{X}}\leq 1$ will ensure that $\|W_\ell\|_{\mathcal{W}_{\ell}}\leq \rho_{\ell}$ with a layerwise radius $\rho_\ell$ for all $\ell\in[L+1]$. 
\tp{Note, that in contrast with the first part on optimization algorithms, which only covers vector and matrix cases, this choice places a product norm over a \emph{collection} of matrices and vectors.}

The $\lmo$-based schemes such as \ref{eq:SCG} decompose layerwise under this norm since
\begin{equation*}
\lmo_{\|\cdot\|_{\mathcal{X}}}(d)=\{\rho_1\lmo_{\|\cdot\|_{\mathcal{W}_1}}(d_1), ..., \rho_{L+1}\lmo_{\|\cdot\|_{\mathcal{W}_{L+1}}}(d_{L+1})\}
\end{equation*} %
\tp{for $d=\{d_1,...,d_{L+1}\}$ where $\lmo_{\|\cdot\|_{\mathcal{D}}}(d):=\argmin_{\|x\|_{\mathcal{D}}\leq 1}\braket{d,x}$.  
The layerwise update magnitudes are therefore dictated by the radii, i.e., $\|\lmo_{\|\cdot\|_{\mathcal{W}_{\ell}}}(d_{\ell})\|_{\mathcal{W}_{\ell}} = \rho_{\ell}$ for all $\ell\in[L+1]$, which in turn guarantees that the upper bound in spectral condition (\Cref{def:spectral}) holds for all iterates of \ref{eq:SCG} as long as $\lambda_k$ and $\rho_\ell$ are chosen in a width-independent way (we are free to choose these parameters in practice).
This layerwise decomposition is used in the Scion optimizer, which chooses $\max_\ell \|W_\ell\|_{\mathrm{op_\ell}}$ for some layer dependent operator norms (see \Cref{tbl:operatornorms}) that can all be made to satisfy the upper bounds in \cref{def:spectral} with appropriate scaling. %
This layerwise decomposition is also present the in LARS and LAMB optimizers, which instead choose the norm max-norm over Frobenius norms, $\max_\ell \|W_\ell\|_F$.} %

\begin{table*}
\centering
\caption{\tp{Example operator norms and the associated $\lmo$s of a matrix $A \in \R^{p_\mathrm{out} \times p_\mathrm{in}}$. The reduced SVD is given as $A=U\diag(\sigma) V^\top$, $\sign$ acts elementwise, $\operatorname{col}_j(A):=A_{\cdot,j}$ and $\operatorname{row}_i(A):=A_{i,\cdot}$. This table is not exhaustive.}}
\label{tbl:operatornorms}
\bgroup
\def\arraystretch{1.2}
\resizebox{\textwidth}{!}{
\begin{tabular}{|c|c|c|c|c|}
\hline
& $1 \rightarrow \RMS$ (ColNorm) & $1 \rightarrow \infty$ (Sign) & $\RMS \rightarrow \RMS$ (Spectral) & $\RMS \rightarrow \infty$ (RowNorm) \\
\hline\hline
\textbf{Norm} & $\max_j \tfrac{1}{\sqrt{p_\mathrm{out}}}\|\operatorname{col}_j(A)\|_2$ & $\max_{i,j} |A_{i,j}|$ & $\sqrt{\nicefrac{p_\mathrm{in}}{p_\mathrm{out}}}\|A\|_{\mathcal{S}_{\infty}}$ & $\max_i \sqrt{p_\mathrm{in}}\|\operatorname{row}_i(A)\|_2$ \\
\hline
\textbf{LMO} & $\operatorname{col}_j(A)\mapsto -\sqrt{p_\mathrm{out}}\tfrac{\operatorname{col}_j(A)}{\|\operatorname{col}_j(A)\|_2}$ & $A\mapsto -\sign(A)$ & $A\mapsto-\sqrt{\nicefrac{p_\mathrm{out}}{p_\mathrm{in}}}UV^\top$ & $\operatorname{row}_i(A)\mapsto -\tfrac{1}{\sqrt{p_\mathrm{in}}}\tfrac{\operatorname{row}_i(A)}{\|\operatorname{row}_i(A)\|_2}$ \\
\hline
\end{tabular}
}
\egroup
\end{table*}

This max-norm over layers has the benefit of decomposing the $\lmo$ computation.
The steepest descent methods \ref{eq:SSD} on the other hand, require computing the dual norm $\|d\|_{\mathcal{X}_*}=\sum_\ell \|d_\ell\|_{\mathcal{W}_*}$, which entails multiplying each layer's update by the global sum of the layerwise norms.

More importantly, the choice of the maximum norm over layers leads to an update of \emph{all} layers at a given iteration.
This is in contrast with, e.g., the choice of the $\ell_1$-norm made in the otherwise prescient work by Flynn \cite{flynn2017duality} on dualization of neural networks, which leads to a (block) coordinate-wise update rule which modifies only one layer per step.
The choice of maximum norm is one of the key design choices in the modular norm \cite{large2024scalable,bernstein2024modular}, which uses the max-norm to generate layerwise scalings of the stepsize that can lead to hyperparameter transfer. In \cite{large2024scalable}, the modular norm is formed exactly by taking the max-norm over layerwise norms. However, its main practical use was for layerwise stepsize normalization, which corresponds to assigning a scaled Euclidean norm to each layer.

\paragraph*{Layerwise operator norms}
\tp{
It turns out that constraining the weight matrices in $\mathrm{RMS}\to\mathrm{RMS}$ norm is also a way to ensure that the layers are Lipschitz in the same norm, under certain assumptions on the input and the loss function computed on the output \cite{large2024scalable}}. %
This operator norm perspective is the key idea behind recent works that led to Muon and Scion \cite{bernstein2024modular,pethick2025trainingdeeplearningmodels}. In \cite{bernstein2024modular} it was proposed to assign operator norms to each layer and compute the $\lmo$ with respect to them (although no consideration was given to stochasticity or boundedness of the parameters). 
In \cite{pethick2025trainingdeeplearningmodels}, it was shown that same principle of assigning operator norms can be extended and used with the stochastic conditional gradient algorithm, to guarantee bounded parameters.
If we move beyond assigning the $\mathrm{RMS}$ norm to the input and output space, we can constrain our weight matrix in a different operator norm and ensure the resulting layer is Lipschitz with respect to the new norms. %
Specifically, \cite{bernstein2024modular} proposed the column-wise RMS norm (ColNorm) for the input layer, while \cite{pethick2025trainingdeeplearningmodels} proposed the row-wise RMS norm (RowNorm) for the output layer (\textit{cf}., \Cref{tbl:operatornorms}).

\paragraph*{Revisiting constrained optimization}
The Lipschitz condition naturally requires, as an assumption, that the parameters themselves are contained within a norm-ball described above.
This provides additional motivation for the norm-ball constrained optimization formulation \eqref{eq:min:constrained} from the optimization section.
Due to the max-structure of the norm choice in \eqref{eq:maxnorm}, the stochastic conditional gradient method (SCG), a.k.a. stochastic Frank-Wolfe, decomposes layerwise,
\begin{equation*}
W^{k+1}_\ell = (1-s_k\lambda) W^k_\ell + s_k\lambda \lmo_{\|\cdot\|_{\mathcal{W}_\ell}\le \rho_\ell}(d^k_\ell)   = (1-s_k\lambda) W^k_\ell + s_k\lambda \rho_\ell\lmo_{\|\cdot\|_{\mathcal{W}_\ell}\le 1}(d^k_\ell)
\quad \forall \ell \in [L+1], 
\end{equation*}
where we have decomposed the dual feedback (a.k.a. the gradient estimator) as $d^k = \{d^k_1,..., d^k_{L+1}\}$. %
Note that $\rho_\ell$ is the effective radius of the norm-ball constraint and that it is defined \emph{layerwise}.

In particular, we see that the radius (and thus the regularization) of the last ($(L+1)^{\mathrm{th}}$) layer increases with width, i.e., $\rho_{L+1} = \tfrac{\rho}{\sqrt{p_{L}}}$ for some width-independent scaling factor $\rho$, with the spectral norm choice from \Cref{def:spectral}.
This is in contrast with hidden layers which have $\rho_\ell = \rho\sqrt{\tfrac{p_\ell}{p_{\ell-1}}}$, which remains a constant in the limit.

\begin{figure}
\centering
\includegraphics[width=0.30225\linewidth]{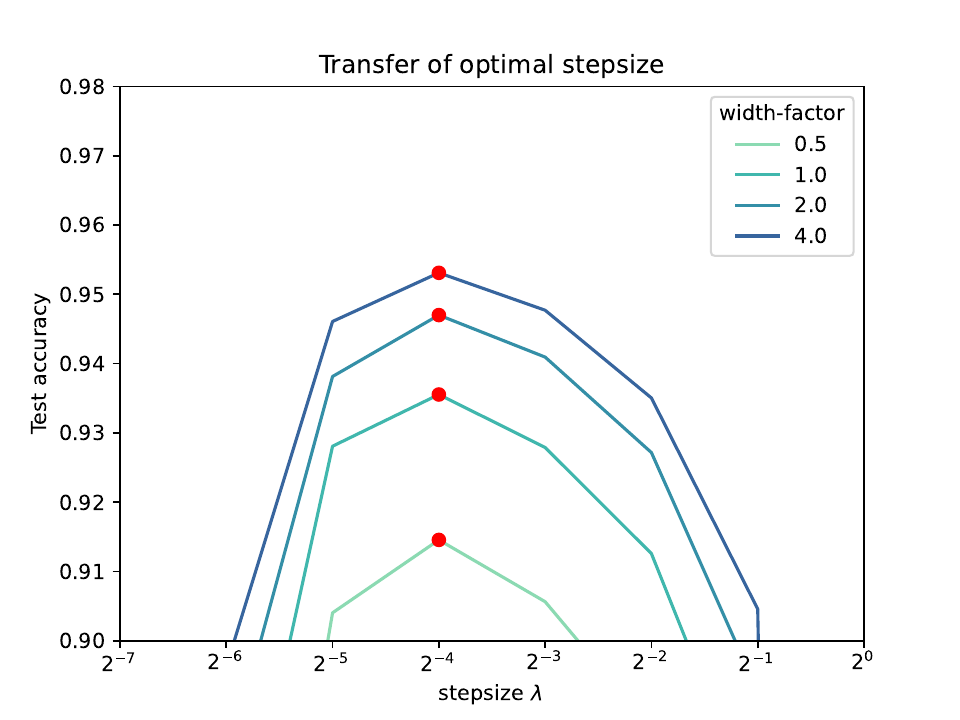}
\includegraphics[width=0.30225\linewidth]{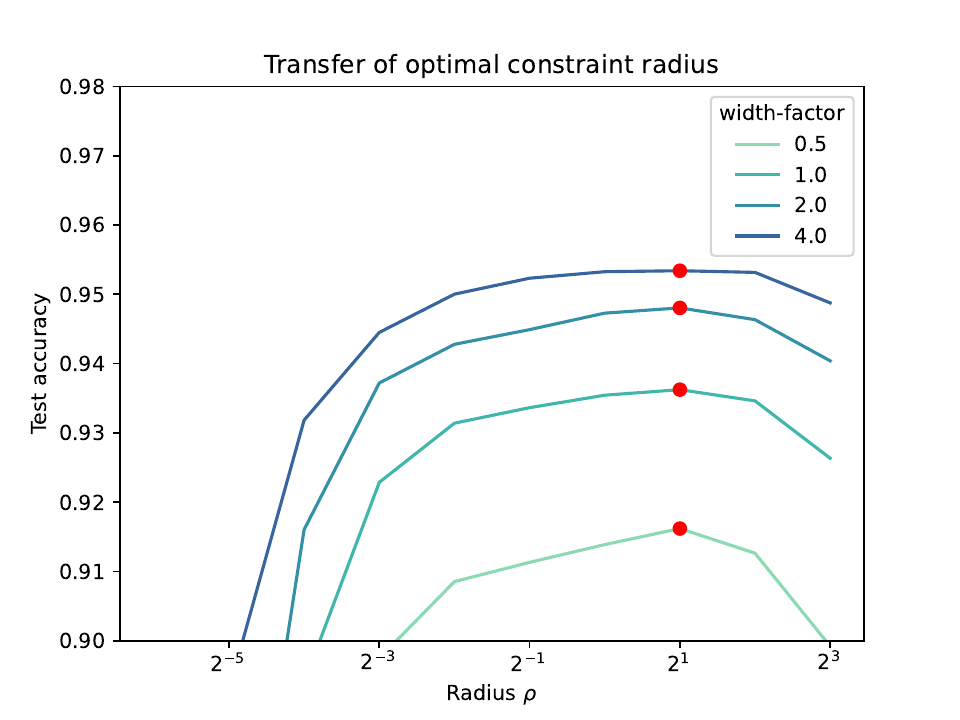}
\includegraphics[width=0.3775\linewidth]{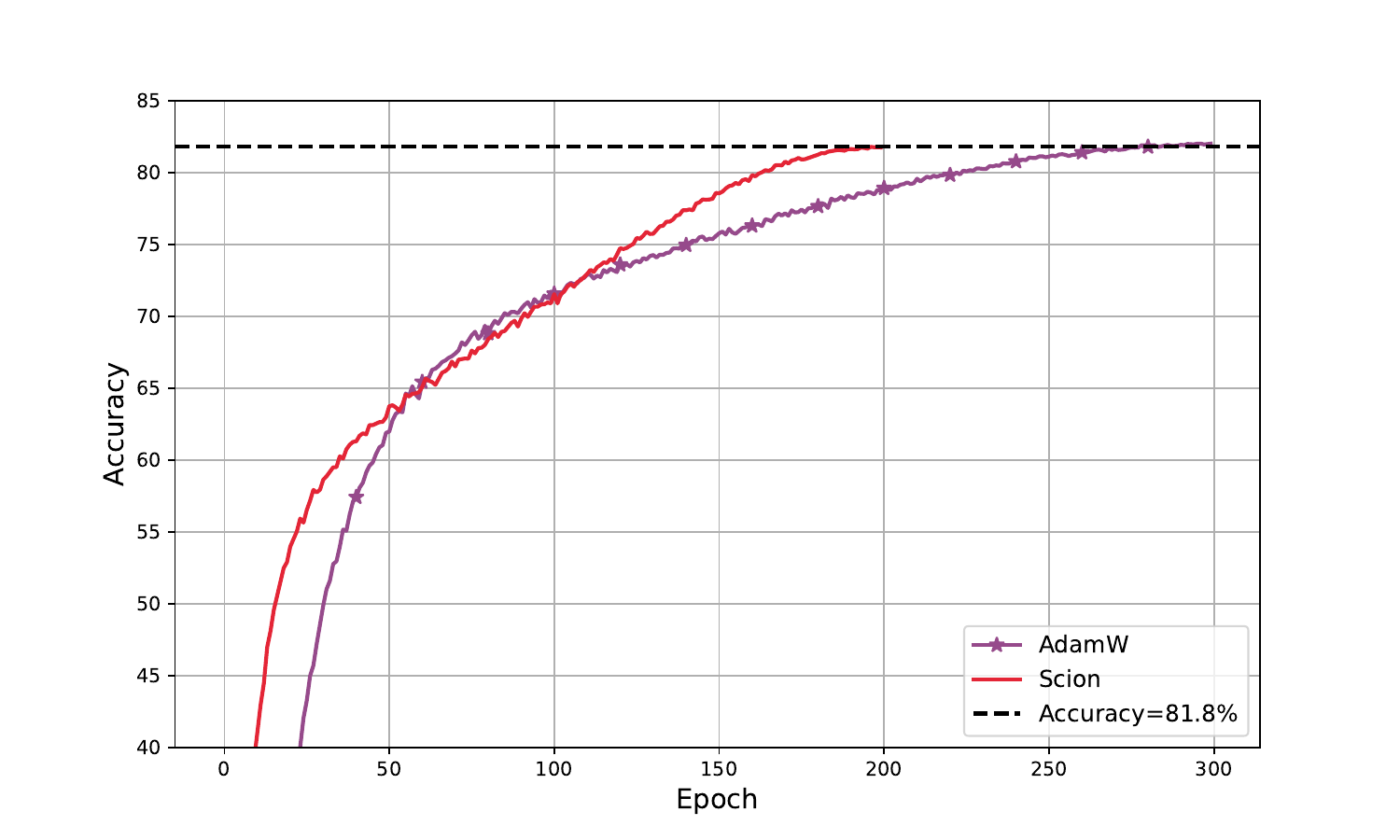}
\caption{The optimal stepsize and constraint radius transfers, when the norm constrained is picked appropriately layerwise for a CNN (left, middle). On the right, we see improved performance when training a vision transformer to classify ImageNet using a geometry-aware optimizer like Scion (plot from \cite{pethick2025trainingdeeplearningmodels})}
\label{fig:transfer}
\end{figure}

We are now going to think about the Lipschitz conditions as a non-asymptotic version of $\mu$P. Empirically, this $\mu$P scaling of the spectral norm turns out to lead to hyperparameter transfer as illustrated in \Cref{fig:transfer} on an image classification task on the CIFAR10 dataset using a convolutional neural network (CNN).
What can be observed is that both the stepsize $\lambda$ and the radius scaling $\rho$ transfer across width.
If the radius (a.k.a. the inverse of the weight decay parameter) was kept the same across layers and the stepsize instead varied over layers, then transfer should not be expected for the weight decay.

\paragraph*{Limitations}
The Lipschitz condition complements the infinite limit results, by, on the one hand, providing non-asymptotic results but, on the other hand, being unable to provide guarantees for feature learning.
The second issue arises since Lipschitz conditions inherently only require an argument regarding upper bounds, which we already saw manifest itself when making the choice of the max norm over layers.
Without a matching lower bound, we cannot guarantee that the lazy training regime is avoided.

\looseness=-1
It should also be noted that the Lipschitz conditions might be loose even in terms of their constants (ignoring their width dependency).
In practice, this means that the computed constants are not used to determine Lipschitz-dependent hyperparameters, which are instead chosen through hyperparameter search.

\section*{Outlook}\label{sec:conclusion}
We have covered modern optimizers for deep learning, their geometries, and how to make them agnostic to the size of the model.
Many questions remain open.
While we have primarily focused on the invariance to width, there is the much broader question of how to scale optimally and predictably:
How do we generally choose, e.g. the number of iterations, batch size, stepsize, model width, and model depth as a function of our compute budget?
Embracing the non-Euclidean perspective, how can we build even faster algorithms?
How can we refine the constrained choice for a broad class of neural network architectures?
At massive scale, system-level concerns become the main bottleneck.
For example, how to distribute the spectral $\lmo$ becomes a challenge, although research in this direction is active \cite{ahn2025dion,amsel2025polar,lau2025polargrad}.
These are exciting questions that will require cross-disciplinary thinking---we hope that with this review article we have managed to pique the interest of the signal processing community.

---------------------------------------------------
\section*{Acknowledgments}

We thank Jeremy Bernstein, Saleh Soltan, and Abhishek Kumar for helpful discussion.
This work was supported as part of the Swiss AI Initiative by a grant from the Swiss National Supercomputing Centre (CSCS) under project ID a06 on Alps.
This work was supported by the Swiss National Science Foundation (SNSF) under grant number 200021\_205011. 
This work was supported by Hasler Foundation Program: Hasler Responsible AI (project number 21043).
Research was sponsored by the Army Research Office and was accomplished under Grant Number W911NF-24-1-0048.
ASF was supported by a public grant from the Fondation Mathématique Jacques Hadamard.
This work was supported by the Gatsby Charitable Foundation, GAT 3850 and GAT 4058.

---------------------------------------------------

All code is available at \url{https://github.com/LIONS-EPFL/scion}

---------------------------------------------------

\bibliographystyle{IEEEtran}
\bibliography{refs.bib,lions-master.bib,bibliography-l2s.bib}

\end{document}